\documentclass[journal]{IEEEtran}

\ifCLASSINFOpdf
 \else
\fi

\usepackage[dvipsnames]{xcolor}
\usepackage{amsmath}
\usepackage{amssymb}
\usepackage{graphicx}
\usepackage{comment}
\usepackage{tabularray}
\usepackage{adjustbox}
\usepackage{cite}
\usepackage{url}
\usepackage{multirow}
\usepackage{multicol}

\newtheorem{definition}{Definition}
\newtheorem{proposition}{Proposition}

\begin{document}

\title{Privacy Preserving Prompt Engineering: A Survey}
\author{\IEEEauthorblockN{Kennedy Edemacu, Xintao Wu}\\
\IEEEauthorblockA{\textit{Department of Electrical Engineering and Computer Science} \\
\textit{University of Arkansas}\\
\{kedemacu, xintaowu\}@uark.edu}
}
\maketitle

\begin{abstract}
Pre-trained language models (PLMs) have demonstrated significant proficiency in solving a wide range of general natural language processing (NLP) tasks. Researchers have observed a direct correlation between the performance of these models and their sizes. As a result, the sizes of these models have notably expanded in recent years, persuading researchers to adopt the term large language models (LLMs) to characterize the larger-sized PLMs. The size expansion comes with a distinct capability called in-context learning (ICL), which represents a special form of prompting and allows the models to be utilized through the presentation of demonstration examples without modifications to the model parameters.
Although interesting, privacy concerns have become a major obstacle in its widespread usage. Multiple studies have examined the privacy risks linked to ICL and prompting in general, and have devised techniques to alleviate these risks. Thus, there is a necessity to organize these mitigation techniques for the benefit of the community. This survey provides a systematic overview of the privacy protection methods employed during ICL and prompting in general. We review, analyze, and compare different methods under this paradigm. Furthermore, we provide a summary of the resources accessible for the development of these frameworks. Finally, we discuss the limitations of these
frameworks and offer a detailed examination of the promising
areas that necessitate further exploration.
\end{abstract}

\begin{IEEEkeywords}
Pre-trained language models, large language models, privacy protection, prompting, in-context learning.
\end{IEEEkeywords}

\IEEEpeerreviewmaketitle

\section{Introduction}

\begin{figure*}[htp]
    \centering
    \includegraphics[width=18cm]{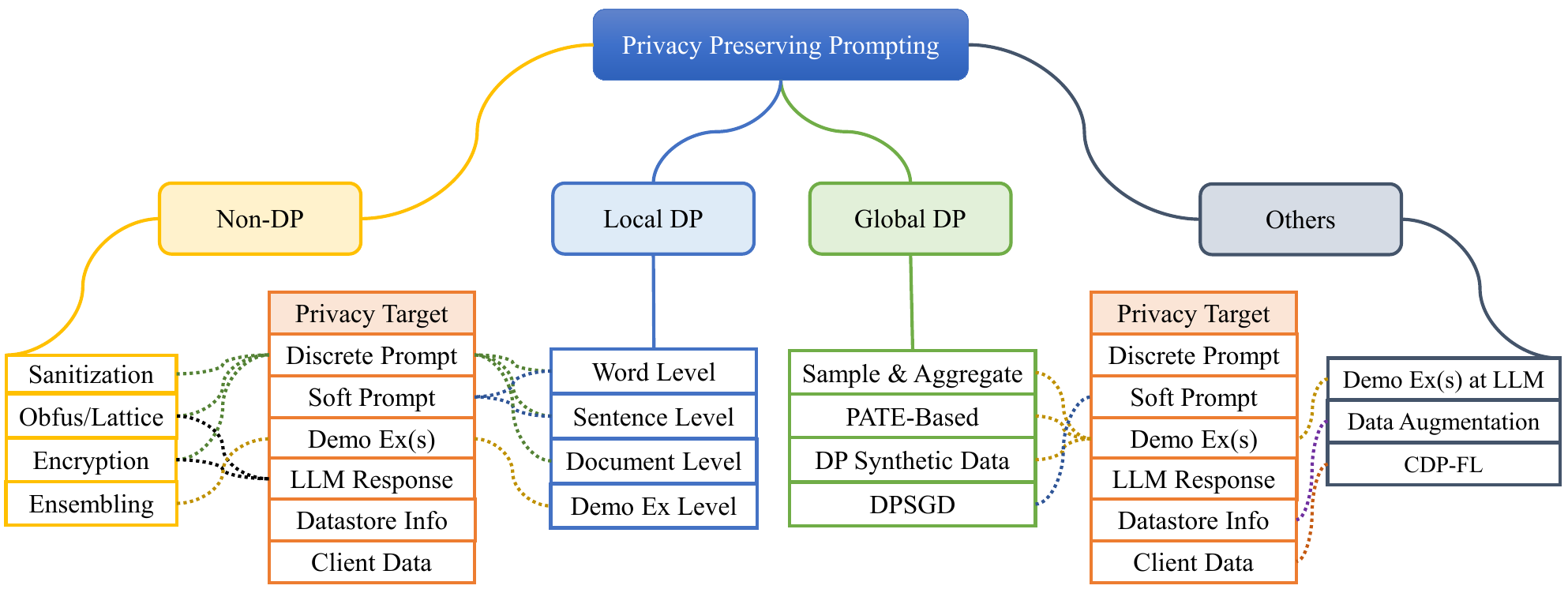}
    \caption{The layout of privacy mechanisms employed for privacy-preserving prompting. Each privacy mechanism protects at least one privacy target. We elaborate on this by creating links between the mechanisms and the privacy targets. Demo Ex(s) denotes demonstration example(s), Obfus denotes obfuscation, and CDP-FL denotes client data protection via federated learning.}
    \label{fig:current_progress}
\end{figure*}

\IEEEPARstart{T}{he} recent advancements in pre-trained language models (PLMs) have demonstrated significant capabilities across a wide array of natural language processing (NLP) tasks such as text classification, question answering, sentiment analysis, information retrieval, summarization, etc \cite{brown2020, liang2022, yu2023, zhang2022}. A number of these models have recently been introduced and are consistently gaining considerable popularity. For example, the number of users for OpenAI’s ChatGPT \cite{openai2023} has surpassed 180 million \cite{duarte2023}. Examples of other common advanced models are Meta's LLaMA \cite{touvron2023}, OPT \cite{zhang2023opt}, OPT-IML \cite{iyer2022opt}, BigScience's BLOOM \cite{workshop2022bloom}, BLOOMZ \cite{muennighoff2022crosslingual}, Databricks' Dolly \cite{conover2023}, etc.

Generally, these models are huge with parameter sizes of hundreds of billions, and require enormous amounts of computational resources for their training and storage. The term \textit{large language models (LLMs)} is employed to delineate these large-sized PLMs \cite{shanahan2024, wei2022, taylor2022}.  Furthermore, these models are primarily pre-trained using diverse open-text resources sourced from the web, books, Wikipedia, etc. For the rest of this work, we shall use the terms PLMs and LLMs interchangeably. While general-purpose LLMs have proven adept at comprehending and solving general NLP tasks, they occasionally demand greater depth and nuance in addressing domain-specific tasks and adapting to specific objectives. Fine-tuning techniques that adjust learnable model parameters have been proposed to tailor the models for domain-specific downstream tasks and to adapt them to specific goals \cite{hu2023llm, hu2021, li2021}. However, challenges such as high computational resource demand, risks of overfitting, concerns about catastrophic forgetting, and model stability are commonly associated with the process of model fine-tuning \cite{zheng2023}. Consequently, caution must be exercised when performing fine-tuning. Other model adaptation techniques include, instruction tuning, prompt tuning, alignment tuning, etc \cite{zhao2023}.

An emerging capability of LLMs is known as \textit{prompting}. Through prompting, an LLM can generate anticipated outputs for a given query when provided with natural language instructions and/or demonstration examples, without necessitating updates to the model parameters. The simplest type of prompt is a direct prompt (also known as zero-shot) where users phrase the instruction as a question and provide no examples to the LLM. In-context learning (ICL) is another form of prompting proposed along with GPT-3 \cite{brown2020} and includes a few demonstration examples in the prompt. ICL serves as an efficient and effective method for leveraging pre-trained or adapted LLMs to address a variety of downstream tasks without the need to modify the model parameters for each task. 

However, the aforementioned LLM utilization technique invariably entails the use of data that may be deemed private and harbor sensitive information. For example, consider using ICL to predict if an individual earns at least \$50,000 in a year. To help the LLM form a context and make a better prediction, it is prompted with a demonstration example that might contain sensitive information such as age, salary, and SSN. 
Sensitive information of this nature could potentially be accessed by either an untrusted LLM server or an adversarial entity capable of bypassing the API provided by the LLM service provider. In addition, due to vulnerabilities in Redis client open-source library \cite{chatgptleak}, ChatGPT leaked users' chat history.
While privacy challenges such as training or fine-tuning data memorization, and their subsequent recovery through model inversion and membership inference attacks \cite{zhang2022text, morris2023, wang2023decodingtrust}
 have been noticed \cite{duan2023}, they are fundamentally distinct from the privacy challenges posed by ICL.
 Therefore, addressing the privacy challenges associated with ICL specifically, as well as prompting in general, is a matter of urgency. 

We focus on privacy concerns when utilizing LLMs with users' sensitive data incorporated into prompts.
Progresses dedicated to mitigating these privacy challenges have been made \cite{duan2023, tang2023, hong2023, duan2023flocks, wutschitz2023}. These studies have adopted various privacy protection mechanisms such as differential privacy (DP), sanitization, lattice, encryption, ensembling, etc in designing their privacy protection frameworks. 
In this survey, we review the generic approaches that preserve privacy during prompting in general. We characterize them by their privacy models, summarize them, and draw links between them. Figure \ref{fig:current_progress} provides a concise overview of the categorization of privacy-preserving prompting. We classify most privacy frameworks into four main categories: non-DP, local DP, global DP, and other scenarios. Within each category, we specify privacy mechanisms/definitions and emphasize their respective privacy objectives. 

From a broader perspective, privacy protection for LLMs can be categorized into two families: protecting personal privacy in the pre-training or fine-tuning corpus, and protecting sensitive information in a user's prompt input data.
There have been a number of surveys focusing on privacy issues in PLMs. Our primary focus in this paper is on the privacy protection aspects during ICL in specific and prompting in general. We now present similar existing surveys in the literature. 
Hu \textit{et al.} \cite{hu2023} specifically explored privacy preservation in NLP using differential privacy. 
Their work mainly focused on how to privately train and release a language model without leaking information about training and fine-tuning data.
Yao \textit{et al.} \cite{yao2023} investigated how LLMs could alter the cybersecurity world. They explored the merits, risks, and vulnerabilities associated with LLM usage. Neel and Chang \cite{neel2023} presented privacy issues in LLMs from training to inference, with limited coverage of the prompting methods. Sun \textit{et al.} \cite{sun2024trustllm} thoroughly investigated the trustworthiness of LLMs across different dimensions including truthfulness, safety, fairness, robustness, privacy, and machine ethics. However, the privacy aspect centers on assessing privacy awareness within LLMs and the potential disclosure of private information from the training dataset in the responses generated by LLMs. Das \textit{et al.} \cite{das2024security} examined the security and privacy challenges associated with LLMs, encompassing considerations for both training data and users. The study evaluated the vulnerabilities of LLMs, explored emerging security and privacy threats targeting LLMs, and provided a review of potential defense mechanisms, although with limited coverage for prompting privacy.
In contrast, our work holistically focuses on a review of the literature on privacy protection in LLM usage with prompting methods. To the best of our knowledge, we are the first to systematically organize such literature.

\textit{Paper organization:} We organize the rest of the paper as follows: Section \ref{sec:formal_description} presents the preliminary section, introducing language models and privacy models. Sections \ref{sec:privacy_prompt_tuning} presents the review of the efforts for privacy-preserving prompting. In Section \ref{sec:resources}, we present the available resources for privacy-preserving prompting. Section \ref{sec:future_prospects} presents the limitations of the existing frameworks and future prospects. We conclude the paper in Section \ref{sec:conclusion}.

\section{Preliminaries}\label{sec:formal_description}
\subsection{Language Models}
\subsubsection{Pre-trained Language Models}
With the abundance of extensive unlabelled corpora and the rise of Transformers \cite{vaswani2017}, the research community has crafted universal pre-trained language models (PLMs) employing self-supervised learning methods \cite{liu2021}. The term large language models (LLMs) is introduced to distinguish PLMs characterized by vast parameter sizes, typically comprising tens or hundreds of billions of parameters. Through pre-training on extensive corpora, LLMs develop the capability to comprehend and generate natural languages proficiently.

Generally, these language models are designed to output the probability distribution of a token sequence \cite{zhao2023}. Given a text sequence $s=\langle w_1, w_2,\cdots,w_n \rangle$, where $w_i\in\mathcal{V}$ and $\mathcal{V}$ denotes the vocabulary space. The likelihood of $s$ using the chain rule of probability is $\Pr(s)=\prod_{i=1}^{n}\Pr(w_i|w_{<i})$. An autoregressive language model takes a sequence of tokens $\langle w_1,w_2,\cdots,w_{i-1} \rangle$ as input and outputs a probability distribution for the next token $w_i$ as $\Pr(w_i|w_{<i})$. Choosing the next token can be achieved through various techniques, including greedy search, beam search, top-$k$ sampling, nucleus sampling, etc. Several surveys \cite{raffel2020, zhao2023, liu2023, qui2020, han2021, doddapaneni2021, zheng2023} have extensively covered the technical details of these LLMs.

Most LLMs, e.g., GPT4 from OpenAI and Germini from Google, only support API-based accesses.  Users submit their prompts via LLMs' standard APIs that provide the input/output functions of the models. A limited number of LLMs such as LLaMA from Meta are open source and provide users all details of system architecture, parameters, and code. For these white-box LLMs, users can retrain or fine-tune models locally for their application task. While pre-training equips LLMs with the capacity to solve diverse NLP tasks, research indicates that their proficiency can be tailored to specific tasks with techniques such as fine-tuning, prompt tuning, instruction tuning, alignment tuning, etc \cite{zheng2023, zhao2023}. Our emphasis lies not in tuning/adaptation, but rather in leveraging the LLMs post pre-training and/or adaptation using prompting.

\subsubsection{Prompting}
A major approach to efficiently and effectively utilize LLMs to solve specific downstream tasks is through designing suitable prompting strategies. A lot can be achieved with well-designed prompts. Manually crafting prompts can be time-consuming and prone to errors. Improperly constructed prompts can lead to poor performance. As a result, a series of efforts have been aimed at automating the optimization of \textit{discrete/hard} and \textit{continuous/soft} prompts \cite{li2021, shin2020}. A \textbf{discrete prompt} typically consists of a sequence of natural language text and can be formulated as:
\begin{equation}\label{eq:prompting}
LLM(prompt)\xrightarrow{} response
\end{equation}
where \textit{prompt}  encompasses elements such as task description, input data, contextual information, prompt style, etc. Those elements are essential for guiding LLMs to generate an appropriate \textit{response}. Formally, we can describe \textit{prompt} as a token sequence $s= \langle w_1, w_2,\cdots, w_l \rangle$ with length $l$. The optimization process in discrete prompts searches for prompts in the discrete text space. Common categories of discrete prompt optimization methods include gradient-based, reinforcement learning (RL)-based, edit-based, and LLM-based approaches \cite{zhao2023}. 

A \textbf{continuous prompt} consists of a continuous set of task-specific embeddings. Note that all discrete input tokens to LLMs are internally transformed into continuous input embeddings that the LLM then processes. In the white-box setting, LLMs further support the user's access via submitting the embeddings of the prompt.  The formulation can be modified as: 
\begin{equation}\label{eq:softprompting}
LLM(embeddings)\xrightarrow{} response
\end{equation}

We can choose a word embedding model $\phi : \mathcal{V} \mapsto \mathbb{R}^d$ to derive the prompt embeddings at the token level, $\langle \phi(w_1), \phi(w_2),\cdots, \phi(w_l) \rangle$,  or use an encoder $\mathbf{r}= \text{Enc}(s)$ that returns a vector representation $\mathbf{r}$ for the whole prompt.  The continuous prompts can also be considered trainable parameters and can be learned with optimization strategies such as employing supervised learning to minimize cross-entropy loss using sufficient downstream task data, and engaging in prompt-based transfer learning \cite{zhao2023}.

\subsubsection{In-Context Learning (ICL)}
\begin{figure}[htp]
    \centering
    \includegraphics[width=\columnwidth]{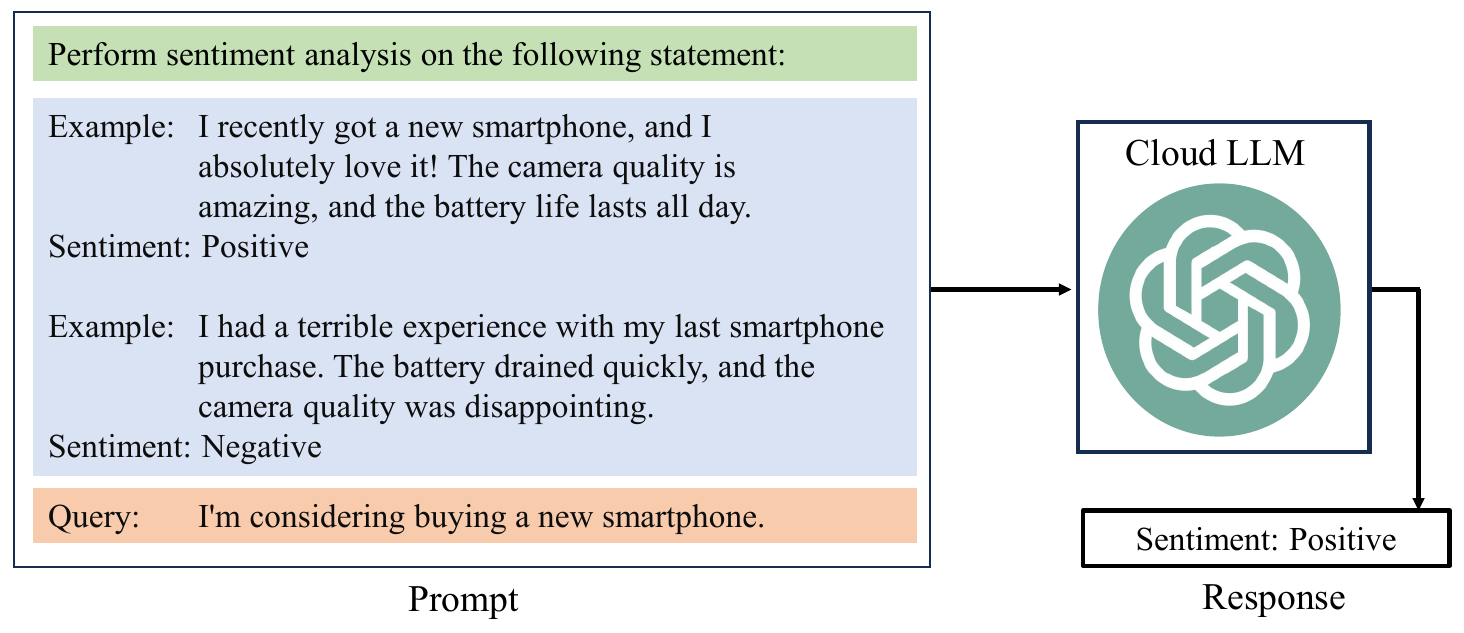}
    \caption{An illustration of ICL. In ICL, the prompt consists of a task description (light green), demonstration examples (light blue), and a query (light orange).}
    \label{fig:icl}
\end{figure}

ICL exemplifies a form of prompting method \cite{brown2020, dong2022} and has become a new learning paradigm where LLMs make predictions only based on contexts presented through a few demonstration examples. In ICL, prompts are composed of task descriptions and/or demonstration examples presented in natural language text. The key idea of in-context learning is to learn from analogy. An illustration of ICL is presented in Figure \ref{fig:icl}. 

A query is concatenated to the demonstration examples usually written in natural language templates. Once done, the combination of the task description, demonstration examples, and the query are submitted to a remote LLM (mostly hosted in the cloud). The LLM is expected to learn the patterns hidden in the demonstration and make the prediction directly without conducting parameter updates. This is different from supervised learning requiring a training stage that often uses backward gradients to update model parameters.  Using appropriate demonstration examples, the LLM can be steered to correctly answer the query. 
Let $D_k = \{(x_1, y_1),\cdots,(x_k, y_k)\}$ denote a set of $k$ demonstration examples. We denote $g(x_k, y_k)$  a prompt function (e.g., a template) that transforms the $k$-th demonstration example into natural language text. For the task description $inst$, a set of demonstration examples $D_k$, and a test query $x_{k+1}$, the prediction output $\hat{y}_{k+1}$ from LLMs can be formally formulated as \cite{zhao2023}:
\begin{equation}\label{eq:icl}
LLM(inst, \underbrace{g(x_1, y_1),\cdots,g(x_k, y_k)}_\text{demonstration examples}, g(\underbrace{x_{k+1}}_\text{query}, \underbrace{\_\_\_\__{ }}_\text{answer}))\xrightarrow{} \hat{y}_{k+1}
\end{equation}
where the true answer $y_{y+1}$ is left as a blank to be generated by the LLM. The advantage of this paradigm is that a single trained model can be used to efficiently solve a myriad of downstream tasks in an unsupervised manner \cite{liu2023}. The performance of ICL depends on the appropriate construction of prompts, in particular, how to select those $k$ demonstration examples. ICL has an inherent connection with instruction tuning as both utilize natural language to construct the task or instances. The main difference is that instruction tuning needs to fine-tune LLMs for adaption whereas ICL only prompts LLMs for utilization. We suggest the readers refer to the survey paper \cite{dong2022} for a comprehensive review of ICL. 

Improving ICL can involve implementing strategies such as chain-of-thoughts \cite{wei2022} that incorporates intermediate reasoning steps within prompts, and planning \cite{zhou2022} which decomposes complex tasks into sub-tasks and devises plans to address them one by one. These strategies prove beneficial for models equipped with extensive knowledge about the given task. Users may as well enrich their prompts with private external datastores. Such a strategy is referred to as database augmentation and allows the introduction of knowledge from external sources such as training corpus, external data, unsupervised data, etc \cite{borgeaud2022, yasunaga2022, wang2023}. This can help improve the context base within the prompt. Generally, a database augmentation model consists of a retriever and a generator \cite{zheng2023}. Here, the retriever can retrieve information from external private sources based on a user query, and the generator combines the retrieved information with instructions to form a prompt.

\subsection{Privacy Models}
\subsubsection{Differential Privacy (DP)}
Differential privacy has been a de facto standard for preserving privacy in a myriad of machine learning tasks. 
There are two popular variants of DP, \textit{global} DP and \textit{local} DP. Global DP (GDP)\cite{dwork2006} assumes the existence of a trusted data curator. The curator has access to all individuals' raw data and processes it using a randomized algorithm $\mathcal{A}$. 

\begin{definition}[Global DP]
\label{def:gdp}
 A randomized algorithm $\mathcal{A}$ satisfies $(\varepsilon, \delta)$-GDP, if for any two neighboring datasets $D$ and $D^\prime$ which differ one single record, and for any output $\mathcal{O}\subseteq \text{Range}(\mathcal{A})$, the following holds:
\begin{equation*}
    \Pr(\mathcal{A}(D) \in \mathcal{O}) \leq exp(\varepsilon)\Pr(\mathcal{A}(D^\prime) \in \mathcal{O}) + \delta
\end{equation*}
\end{definition}
where \noindent $\varepsilon$ is the privacy budget and controls the level of privacy guarantee. The smaller the value, the stronger the privacy guarantee (i.e., more noise is added) and vice-versa. $\delta$ is a small error probability. If $\delta = 0$, $\mathcal{A}$ is $\varepsilon$-DP.  Global DP guarantees output of an algorithm be insensitive to the presence or absence of  one record in a dataset.

Local DP (LDP) \cite{kasiviswanathan2011} does not require the existence of a trusted central data curator. Individuals locally perturb their data using a randomized algorithm $\mathcal{A}$ before sending them to the curator for analysis. 

\begin{definition}[Local DP]
\label{def:ldp}
  A randomized algorithm $\mathcal{A}$ satisfies $\varepsilon$-LDP, if for any two inputs $x, x^\prime \in \mathcal{X}$, and for any output $\mathcal{O}\subseteq \text{Range}(\mathcal{A})$, the following holds:
\begin{equation*}
    \Pr(\mathcal{A}(x) \in \mathcal{O} ) \leq exp(\varepsilon)\Pr(\mathcal{A}(x^\prime)  \in \mathcal{O})
\end{equation*}
\end{definition}
Different from the global DP, the inequality holds for all elements  $x$ and $x^\prime$ instead of all adjacent pairs of the dataset. 
Essentially, the local DP ensures that an adversary is unable to infer the input values of any target individual from the output values obtained. To achieve the LDP in statistical analysis, mechanisms such as randomized response, histogram encoding, unary encoding, or local hashing are applied during the collection of user data that are categorical in nature~\cite{DBLP:journals/sensors/WangZFY20}. 

Local DP is a strong privacy notion because it aims homogeneous protection over all input pairs, i.e., no matter how unrelated two inputs $x$ and $x'$ are, their output distributions must be similar. To improve utility, metric LDP \cite{alvim2018local} was proposed where the indistinguishability of output distributions is further scaled by the distance between the respective inputs.  

\begin{definition}[Metric LDP]
\label{def:mldp}
  A randomized algorithm $\mathcal{A}$ satisfies $\varepsilon$-Metric LDP, if for any two inputs $x, x^\prime \in \mathcal{X}$, and for any output $\mathcal{O}\subseteq \text{Range}(\mathcal{A})$, the following holds:
\begin{equation*}
    \Pr(\mathcal{A}(x)  \in \mathcal{O}) \leq exp(\varepsilon \cdot d(x,x'))\Pr(\mathcal{A}(x^\prime)  \in \mathcal{O}) 
\end{equation*}
where $d(\cdot,\cdot)$ is a distance metric. When  $d(\cdot,\cdot)=1$, metric LDP is equivalent to LDP.  
\end{definition}

\subsubsection{Differential Privacy Properties}

Several important properties of differentially private mechanisms arise from the above DP definitions \cite{dwork2014}. In the following sections, we describe those commonly used in NLP. 

\textbf{(a) Post-processing property:} Post-processing an output of a differentially private algorithm cannot reverse its privacy protection. In other words, the output of $(\varepsilon, \delta)$-DP mechanism  remains $(\varepsilon, \delta)$-DP after post-processing. 

\begin{proposition}[Post-processing property]
\label{prop:post-processing}
Let $\mathcal{A}(\mathcal{X})$ satisfy $(\varepsilon, \delta)$-DP. Then, for any (randomized) algorithm $f$,  $f\circ \mathcal{A}(\mathcal{X})$ satisfies $(\varepsilon, \delta)$-DP.  
\end{proposition}

\textbf{(b) Composition property:} The composition property guarantees DP privacy when releasing multiple outputs of DP mechanisms on the same data.

\begin{proposition}[Composition property]
\label{prop:composition}
Suppose $\mathcal{A}_1, \mathcal{A}_2 \cdots, \mathcal{A}_k$ are each $(\varepsilon_i, \delta_i)$-DP algorithms. An algorithm $\mathcal{A} = \mathcal{A}_1 \circ \mathcal{A}_2 \circ \cdots \circ \mathcal{A}_k$ that runs $\mathcal{A}_i$ sequentially satisfies $(\varepsilon, \delta)$-DP where $\varepsilon = \sum_{i}^{k}\varepsilon_i$ and $\delta = \sum_{i}^{k}\delta_i$. 
\end{proposition}

\textbf{(c) Privacy amplification via subsampling property:}  Subsampling further helps to improve sample secrecy by introducing additional randomness.

\begin{proposition}[Amplification Effect of Sampling~\cite{DBLP:conf/ccs/LiQS12}]
\label{prop:amplify}
Suppose $\mathcal{A}$ is $(\varepsilon, \delta)$-DP and $\mathcal{B}$ is constructed as follows. Given a dataset $D=\{x_1, x_2, \cdots, x_n\}$, first, we create a sub-sampled dataset $D_s$. The probability $x_i \in D_s$ is $q$. Next, we run $\mathcal{A}$ on $D_s$. Then $\mathcal{B}(D) = \mathcal{A}(D_s)$ is $(\Tilde{\varepsilon}, \Tilde{\delta})$-DP where $\Tilde{\varepsilon} = \ln(1+(e^\varepsilon-1)q)$ and $\Tilde{\delta} = q\delta$.
\end{proposition}

\subsubsection{Differential Privacy Mechanisms}

In the following, we introduce several commonly used mechanisms to achieve DP. The mechanisms of achieving differential privacy mainly include the classic approach of adding Laplacian noise \cite{dwork2006calibrating}, the exponential mechanism \cite{mcsherry2007mechanism}, the sample and aggregate framework \cite{Nissim2007}, the Private Aggregation of Teacher Ensembles  framework  \cite{papernot2018}, the functional perturbation approach  \cite{Chaudhuri2011}, and the differentially private stochastic gradient descent \cite{abadi2016}.  

Dwork et al. \cite{dwork2006calibrating} proved using Laplace mechanism can preserve differential privacy by calibrating the standard deviation of the noise according to the sensitivity of the query function.  The sensitivity measures the maximum possible change in the function's output when one record in the dataset changes.
	
\begin{proposition}[Laplace Mechanism]
\label{def:laplace}
Given a dataset $D$ and a query $f$, a mechanism $\mathcal{A}(D)=f(D)+\boldsymbol{\eta}$ satisfies $\varepsilon$-DP, where $\boldsymbol{\eta}$ is a random vector drawn from $Lap(S_f(D)/\varepsilon)$ where the sensitivity $S_f(D)$ is defined as $S_f(D)=\max_{D,D'} ||f(D)-f(D')||_1$.
\end{proposition}

McSherry and Talwar \cite{mcsherry2007mechanism} proposed the exponential mechanism to guarantee differential privacy in non-numeric sensitive queries by sampling according to a mapping function instead of adding noise.  
For a given dataset $D$ and privacy budget $\varepsilon$, the quality function induces a probability distribution over the output domain, from which the outcomes are exponentially chosen. It favors higher scoring classes, while guaranteeing $\varepsilon$-differential privacy.
	
\begin{proposition}[Exponential Mechanism]
\label{def:exponential}
Given a dataset $D$, Let $q:D\rightarrow R$ be a quality function that scores each output class $r \in R$. The sensitivity of this function is defined as
\begin{equation}
S(q(D,r))=\max_{D,D',r\in R} ||q(D',r)-q(D,r)||_1.
\end{equation}
The exponential mechanism $\mathcal{A}$ randomly selects a potential outcome $r$ based on the following probability, then the mechanism $\mathcal{A}_{q,S(q)}^{\epsilon}(D,R)$ is $\varepsilon$-differentially private:
\begin{equation}
\Pr(r\in R \text{ is selected})\propto \exp\left(\frac{\varepsilon q(D,r)}{2S(q)}\right).
\end{equation}
\end{proposition}

Nissim \textit{et al.} \cite{Nissim2007} introduced the sample and aggregate framework. It calibrates instance-specific noise based on a smooth sensitivity to achieve rigorous differential privacy. The private database is randomly split into multiple partitions. The arbitrary function $f$ is computed exactly, without noise, independently on each partition. The intermediate outcomes are then combined via a differentially private aggregation mechanism, e.g., standard aggregations followed by noise perturbation. Because any single element can affect at most one partition, changing the data of any individual can change at most a single input to the aggregation function. 

Papernot et al. \cite{papernot2018} developed the Private Aggregation of Teacher Ensembles (PATE) framework that incorporates both a private labeled dataset and a public unlabeled dataset and ensures DP by employing a teacher-student knowledge distillation framework. The student model acquires knowledge from the private dataset through knowledge distillation facilitated by the multiple teacher models. Specifically, the private dataset is first randomly divided into $m$ disjoint subsets, each of which is used to train a teacher model. For each record in the public dataset, the label outputs from all teacher models are aggregated. The student model is then trained on the public dataset using the label guidance provided by the aggregated teacher models. To achieve the DP protection of the private dataset, the noisy majority votes as labels are adopted in the classification task. 

Chaudhuri et al. \cite{Chaudhuri2011} proposed an objective perturbation approach by perturbing the objective function which is convex and doubly differentiable. Zhang et al. \cite{Zhang2012} further proposed a functional mechanism to enforce differential privacy on general optimization-based models, such as linear regression and logistic regression.
Abadi \textit{et al.}  \cite{abadi2016} proposed a differentially private stochastic gradient descent (DP-SGD) technique, which has been another popular mechanism to achieve DP in deep learning. The procedure of deep learning model training is to minimize the output of a loss function through numerous stochastic gradient descent (SGD) steps. DP-SGD uses a clipping bound on $l_2$ norm of the gradient from an individual input, aggregates the clipped updates, and then adds Gaussian noise to the aggregate. The clipping truncation controls  the sensitivity of the sum of gradients as the sensitivity of gradients and the scale of the noise would otherwise be unbounded. 
\cite{abadi2016} further proposed a moment accounting mechanism which calculates the aggregate privacy bound when performing SGD for multiple steps. The moments accountant is tailored to the Gaussian mechanism and  computes tighter bounds for the privacy loss compared to the standard composition theorems.

\section{Prompting with Privacy Protection}\label{sec:privacy_prompt_tuning}

\begin{table*}
\begin{center}
    \begin{tabular}{|c|ccccc|}
    \hline

    \hline
        Type & Ref-Year & Definition/Method &Privacy Target & Model Architecture & Downstream Task\\
    \hline

    \hline
        \multirow{10}{*}{Non-DP} & \cite{kan2023}-2023 & \multirow{3}{*}{Sanitization} &\multirow{3}{*}{Prompt (Eq.\ref{eq:prompting})} & SpaCy & Classification, Translation\\
        & \cite{chen2023}-2023 & & &Llama, ChatGPT & Information Extraction\\
        &  \cite{zhang2024cogenesis}-2024 &  &  & GPT-4 & Generation \\
        \cline{2-6}
        &  \cite{duan2023}-2023  & Ensembling & Demo Ex(s) (Eq.\ref{eq:icl}) & GPT-2 & Classification \\
        \cline{2-6}
        & \cite{yao2024}-2024 & Obfuscation/  &\multirow{2}{*} {Prompt \& Response (Eq.\ref{eq:prompting})} & - &Classification\\
        & \cite{zhang2023latticegen}-2023 & Lattice& &OPT & Creative Writing\\
        
        \cline{2-6}
        &  \cite{lin2024}-2024 & \multirow{4}{*}{Encryption}  &\multirow{4}{*}{Prompt \& Response (Eq.\ref{eq:prompting})} & GPT-4, Gemini & Classification, Recommendation, TDA\\
        &  \cite{hou2023ciphergpt}-2023 & & & GPT-2 & Generation\\
        &  \cite{hao2022iron}-2022 & & &BERT & Classification \\
        &  \cite{chen2022thex}-2022 & & & BERT & Classification \\

    \hline
        \multirow{14}{*}{LDP} & \cite{lyu2020towards}-2020  & \multirow{9}{*}{Word Level} & \multirow{8}{*}{Prompt (Eq.\ref{eq:prompting})} & Glove, BERT & Classification\\
        &  \cite{plant2021cape}-2021 & & & BERT & Classification \\    
        &  \cite{chen2023customized}-2023 & & & BERT, ClinicalBERT & Classification\\
        &  \cite{tong2023privinfer}-2023 & & & GPT-4, Vicuna-7B, FastChat& Generation\\
        & \cite{feyisetan2020privacy}-2020 & & &Glove, FastTest & Classification, QA \\
        & \cite{xu2020differentially}-2020 & & & Glove, FastTest &Classification\\
        & \cite{carvalho2023tem}-2023 & & & Glove & Classification\\
        &\cite{yue2021}-2021 & & & BERT, ClinicalBERT & Classification\\
        & \cite{zhou2023textobfuscator}-2023 & & Prompt Embeddings (Eq. \ref{eq:softprompting}) &RoBETa-base&Classification\\
        
        \cline{2-6}        
        & \cite{du2023sanitizing}-2023& \multirow{2}{*}{Sentence Level} &Prompt (Eq.\ref{eq:prompting}) & BERT & Classification\\
        & \cite{du2023}-2023  &  & Prompt Embeddings (Eq. \ref{eq:softprompting}) & BERT & Classification\\

        \cline{2-6}
        &  \cite{utpala2023locally}-2023 & Document Level& Prompt (Eq.\ref{eq:prompting}) & GPT-3.5, T5 Flan-T5, Stablelm& Classification\\
        
        \cline{2-6}
        & \cite{carey2024}-2024  & Demo Ex Level & Demo Ex(s) (Eq.\ref{eq:icl}) & Llama-2-7B, Llama-2-13B & TDA\\
    \hline
        \multirow{13}{*}{GDP} & \cite{tang2023}-2023  & \multirow{2}{*}{Sample \& Aggregate} & \multirow{2}{*}{Demo Ex(s) (Eq.\ref{eq:icl})} & GPT-3 & Classification, Information Extraction\\
        & \cite{hong2023}-2023 & & & Vicuna-7b, GPT-3 & Classification, Summarization, QA\\

        \cline{2-6}
        & \cite{duan2023flocks}-2023  & \multirow{3}{*}{PATE-Based} &\multirow{3}{*}{Demo Ex(s) (Eq.\ref{eq:icl})} & GPT-3, Claude& Classification\\
        & \cite{tian2022seqpate}-2022 & & & GPT-2& Generation\\
        & \cite{li_2023_ICCV}-2023 & & & EVA, FreeMatch& -\\
        \cline{2-6}
        & \cite{yue2022}-2022  & \multirow{7}{*}{DP Synthetic Data} &\multirow{6}{*}{Demo Ex(s) (Eq.\ref{eq:icl})} & GPT-2, RoBETa-base& -\\
        & \cite{flemings2024}-2024 & & & GPT-2, DistilGPT-2& -\\
        & \cite{kurakin2023}-2023 & & & Lamda-8B, GPT-2& -\\
        & \cite{carranza2023}-2023 & & & T5 & - \\
        & \cite{xie2024}-2024 & & & GPT, OPT, Llama-2, Vicuna &-\\
        & \cite{carey2024}-2024 & & & Llama-2-7B, Llama-2-13B & TDA\\
        
        &  \cite{meehan2022sentence}-2022 & & Prompt (Eq. \ref{eq:prompting}) & SBERT & Classification\\

        \cline{2-6}
        & \cite{duan2023flocks}-2023  & DPSGD &Prompt Embeddings (Eq. \ref{eq:softprompting}) & \text{RoBETa}-base & Classification\\

    \hline
        \multirow{3}{*}{Others} & \cite{wu2023privacy}-2023  & Demo Ex(s) at LLM & Demo Ex(s) &  GPT-3, OpenLlama-13B &Classification, QA\\
        \cline{2-6}
        & \cite{huang2023}-2023 & \multirow{3}{*}{Data Augmentation} &\multirow{3}{*}{Datastore} & GPT-2 & Information Extraction\\
        & \cite{wutschitz2023}-2023 & & & GPT-3 &Information Extraction\\
        & \cite{arora2023}-2023 & & & MDR & Multi-hop QA \\
        \cline{2-6}
        & \cite{yang2023efficient}-2023 & \multirow{3}{*}{CDP-FL } & \multirow{3}{*}{Client Data} & ViT-B/16 &  Classification\\
        & \cite{su2024federated}-2024 & & & CLIP, ViT & Classification \\
        & \cite{guo2023promptfl}-2023 & & & CLIP & Recognition, Classification\\
    \hline

    \hline
    \end{tabular}
    \caption{An overview of privacy-preserving prompting. Demo Ex(s) denotes demonstration example(s), CDP-FL denotes client data protection via federated learning, TDA denotes tabular data analysis, and QA denotes question answering.}\label{tab:prompt_eng_dp}
\end{center}
\end{table*}

In this section, we systematically organize the privacy preservation methods employed during prompting. Generally, prompts can leak sensitive information and can be accessed by adversarial entities. Drawing from the formulations outlined in Equations \ref{eq:prompting}, \ref{eq:softprompting}, \ref{eq:icl}, our attention centers on safeguarding the privacy of key components: the prompts (including demonstration examples) and the outputs produced by LLMs. We categorize the noteworthy methods according into Non-DP, local DP, global DP, and other scenarios in Table \ref{tab:prompt_eng_dp}.

\subsection{Non-DP Methods}

\subsubsection{Sanitization Methods}\label{sec:sanitization}

Data sanitization methods aim to protect the privacy of the prompt in Equation \ref{eq:prompting}. In general, data sanitization techniques aim to identify and eliminate sensitive attributes that contain personally identifiable information (PII) from the data \cite{ishihara2023}. Quite often, a machine learning model is employed to perform the identification of PIIs and other sensitive attributes.
Kan \textit{et al.} \cite{kan2023} proposed to use a local LLM to sanitize privacy-sensitive user inputs before using the sanitized texts for prompting a cloud LLM.  
The proposed Privacy-Preserving via Text Sanitization (PP-TS) framework consists of three modules: a pre-processing privacy protection module that conducts de-identification, an LLM invocation module, and a post-processing privacy recovery module that recovers the original sensitive information.
The local LLM is presented with text examples and re-writes requirements in the form of an instruction. This guides the local LLM to sanitize sensitive attributes in the user input. Once the sanitized text is generated, the local LLM checks its reasonability. If it is considered reasonable, then it is used to prompt a cloud LLM. Additionally, the original and sanitized versions of the text are kept locally as a Plaintext-Ciphertext pair. This is used to recover and restore sanitized attributes included in the response returned by the cloud LLM in the post-processing stage. 

Instead of using a single local LLM for sanitization, Chen \textit{et al.} \cite{chen2023}  developed a framework called Hide and Seek (HaS) that comprises two models, hiding private entities (Hide-Model) for anonymization, and seeking private entities (Seek-Model) for de-anonymization.
The training phase uses an LLM through prompt engineering to generate a training dataset. This training dataset is used to train Hide-Model and Seek-Model. During the inference phase, a user-submitted text is anonymized using the Hide-Model stored in a terminal device such as a mobile phone, tablet, laptop, etc. The anonymized text can then be used to prompt a cloud LLM. At the same time, the anonymized text together with the original text are sent to the Seek-Model deployed locally. Similarly, the output returned from the cloud LLM is also fed into the Seek-Model. Finally, the Seek-Model de-anonymizes the output returned by the cloud LLM for user consumption. However, training two LLMs presents multiple challenges ranging from computational issues to model effectiveness, thus making this framework appear infeasible.

Zhang \textit{et al.} \cite{zhang2024cogenesis} introduced another approach, called mixed-scale model collaboration \cite{zhang2024cogenesis} that combines the capabilities of a large model in the cloud with a small model deployed locally. This blends public text with private data to personalize usage, thus providing a logical solution to the privacy challenge. The developed CoGenesis framework has two variants: sketch-based and logit-based variants. CoGenesis operates under the assumption that a user's instruction comprises two sections: a general (privacy-insensitive) section and a personal (privacy-sensitive) section. In sketch-based CoGenesis, a user first prompts the cloud LLM with the general instruction and receives a content sketch as a response. By employing the sketch-then-fill approach, the user integrates the sketch with both the general and personal instruction sections, allowing for content personalization using the small local LLM. The logit-based variant of CoGenesis integrates the logits from both the cloud LLM and the small LLM to determine the subsequent tokens. The foundation of achieving privacy in this framework lies in not exposing the personal section of the instruction to the cloud LLM, but rather utilizing it locally. 

\subsubsection{Ensemble-Only Methods} 

Ensemble-only methods focus on safeguarding the privacy of demonstration examples during ICL outlined in Equation \ref{eq:icl}. Prompted LLMs may incur a high risk of disclosing private information of demonstration examples used in prompts. 
Duan \textit{et al.} \cite{duan2023} first studied this privacy leakage through the lens of membership inference attacks (MIAs). The adversary assumes to have access to the prediction probabilities of each possible target class of the test example. The attacker instantiates the MIA to determine whether his input was part of the demonstration examples by examining the prediction probabilities $y$.  \cite{duan2023} compared the MIA risk of prompted and fine-tuned models and observed that the privacy risk of prompting is significantly higher than fine-tuning. To mitigate the privacy risk exposed by prompt membership, \cite{duan2023} proposed to aggregate the prediction probability vectors over multiple independent prompted models into an ensemble prediction.
In this work, $K$ models are prompted, each with a disjoint subset of the private data. For an input $x_i$, each prompted model generates its own output probability $y^{(k)}_{i}$. To generate the final output,  two ensemble methods, Avg-Ens and Vote-Ens, were developed. In Avg-Ens, the final output $\hat{y}_i$ for the input $x_i$ is generated by computing the average of the raw probability vectors of each of the $K$ prompted models as $\hat{y}_i = \frac{1}{K}\sum_{k=1}^{K}y^{(k)}_{i}$. Meanwhile, Vote-Ens relies on a majority vote of all prompted models to generate the final output. For input $x_i$, each prompted model outputs a token prediction from the vocabulary $\mathcal{V}$ with the highest logit value. Suppose $n_v$ denotes the number of prompted models that predict token $v$ as the output for the input $x_i$, the final output $\hat{y}_i$ is computed as
$\hat{y}_i = \operatorname{arg\, max}_{v\in \mathcal{V}} (n_v)$. 

\subsubsection{Obfuscation/Lattice Methods} 

Obfuscation and lattice methods aim to safeguard the privacy of both the prompt and the response described in Equation \ref{eq:prompting}. In the current user-LLM interaction paradigms, the user provides the LLM server with a prompt and the LLM generates a response, with both the prompt and the generated response being accessible to the untrustworthy LLM server. There exist application scenarios in which both user prompts and the generated contents need obfuscation because both can directly affect the user's decisions.

Yao \textit{et al.} \cite{yao2024} introduced the Instance-Obfuscated Inference (IOI) framework to safeguard both the raw input and output in the black-box inference setting. During inference, IOI obfuscates input instances, ensuring that the raw decision distribution does not disclose any sensitive information, while still enabling the user to recover the true decision. In other words, IOI obfuscates the raw input instance using obfuscators designed to intentionally influence the cloud LLM's inference decision, thereby preventing adversaries from deducing the true decision without knowledge of the resolution method and parameters. Instead of sending the real instance, IOI combines it with dummy instances and utilizes a Privacy-Preserving Representation Generation (PPRG) method \cite{zhou2022textfusion} to transform the plaintext combination into privacy-preserving representation. The cloud LLM generates its response based on this combination. Subsequently, IOI employs its Privacy-Preserving Decision Resolution algorithm to extract the true response from the LLM's returned response. This prevents adversaries including the PLM service provider from discovering the true instance and response. 

Zhang \textit{et al.} \cite{zhang2023latticegen} developed the LatticeGen framework that enables a collaborative generation of tokens between the user and the LLM server. On each time step, the user and server conduct the generation token-by-token cooperatively instead of letting the server alone handle the generation process. Here, the user sends the server $N$ tokens, only one of which is the true token and the rest act as noise. For a round $t$, the true token is denoted as $w^{1}_{t}$ and the noise tokens are denoted as [$w^{2}_{t},\cdots,w^{N}_{t}$]. To prevent the server from knowing the true token, the client permutes the tokens as $[\tilde{w}^{1}_{t},\cdots,\tilde{w}^{N}_{t}]$. Prior to sending them to the server, a linear operation is then performed on the permuted tokens as: 
\begin{equation}\label{eq:lattice}
    \text{linearize}(\tilde{W}^{N}_{T}) = [<bos>] + \text{concat}^{T}_{t=1}([\tilde{w}^{1}_{t},\cdots,\tilde{w}^{N}_{t}])
\end{equation}
where $T$ is lattice length and $<bos>$ denotes the beginning of a sentence token. After receiving the linearized tokens, the server generates the next token for each of the $N$ token options and returns them to the user. The user then performs reverse permutation mapping to obtain the true token from the returned tokens. 

\subsubsection{Encryption  Methods}
Encryption-based methods follow the same principles as obfuscation and lattice methods, aiming to safeguard the privacy of both the prompt and the response in Equation \ref{eq:prompting}. Lin \textit{et al.}  \cite{lin2024} introduced an emoji encryption based framework called EmojiCrypt. Here, a user's sensitive prompt inputs are encrypted into emojis before sending them to the cloud LLM. This attempts to render the original input unreadable to humans while retaining enough information for the LLM to perform effectively. In a way, this method is similar to the sanitization method, instead of replacing sensitive text with other non-sensitive phrases, the emoji encryption replaces them with emojis. 

Hou \textit{et al.} \cite{hou2023ciphergpt} developed the CipherGPT framework for secure two-party inference. Secure inference is a two-party cryptographic protocol running inference to achieve the goal that the server learns nothing about the client's input and the client learns nothing about the model except the final inference results. Generally speaking, the protocol proceeds by having the server and client running the encrypted model over the encrypted input through cryptographic techniques such as homomorphic encryption (HE) and secret sharing.   
CipherGPT introduces a series of novel protocols, including a secure matrix multiplication that is customized for GPT inference, a protocol for securely computing GELU, and a protocol for top-k sampling, to support secure GPT inference. 
Other frameworks such as Iron \cite{hao2022iron} and THE-X \cite{chen2022thex} also employ cryptographic techniques to securely perform inference in transformer-based models. Specifically, Iron adopts HE and secrete sharing to build efficient protocols for matrix multiplication and other non-linear operations such as Softmax, GELU and LayerNorm used in a single transformer LLMs. THE-X replaces complex non-linear operations in transformer-based LLMs with HE-friendly operations. For instance, it replaces GELU with ReLU, and Softmax with a combination of ReLU and polynomials.
Although these techniques can be used for privacy preserving text generation tasks theoretically, practical applications are limited because of high computational and communication costs.

\subsection{Local DP Methods}

\begin{figure*}[htp]
    \centering
    \includegraphics[width=18cm]{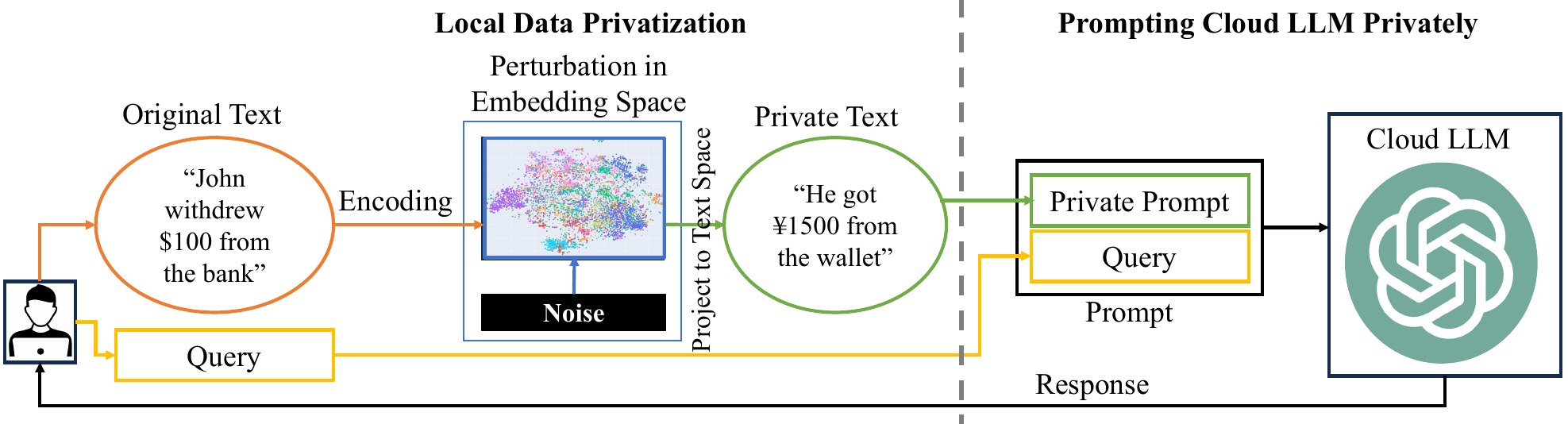}
    \caption{An illustration of privatizing sensitive local data with LDP before using it to prompt a cloud LLM privately.}
    \label{fig:ldp_layout}
\end{figure*}

Simple sanitation techniques discussed in Section \ref{sec:sanitization} fail to provide rigorous privacy protection. LDP mechanisms have been explored to allow users to sanitize their sensitive data locally before sending the sanitized prompt to the untrusted LLM server. Generally speaking, these methods introduce randomness to text by privatizing the embedding vector for each token, word, sentence, or the whole prompt.  The text data is first transformed into a representation vector via embedding methods and some DP mechanism is then applied to privatize representations.  The perturbed vector representations are projected back into the text space to find some appropriate text. Due to the post-processing property of DP, mapping privatized representations back to text also preserves differential privacy. After perturbation, the LDP perturbed prompt is then sent to the remote black-box LLM. The LLM then returns the generation result to the user. Figure \ref{fig:ldp_layout} illustrates this procedure. The original text (``John withdrew \$100 from the bank'') is first encoded to generate its embeddings. LDP noise is then added to the embeddings to generate a perturbed version (``He got \yen1500 from the wallet'') of the text. This private text is then used in the prompt to guide the cloud LLM  to generate a response to the query.

Most LLMs only support black-box access where users submit text prompts via LLMs' standard interfaces. However, a limited number of LLMs also support white-box access. Here, we present LDP methods that can be employed to safeguard the text prompts described in Equation \ref{eq:prompting}, soft prompts described in Equation \ref{eq:softprompting}, and demonstration examples described in Equation \ref{eq:icl}. We categorize them according to their privacy levels as follows.

\subsubsection{Word Level Perturbation}
To achieve privacy protection for each word (or token) in the text prompt, Lyu \textit{et al.} \cite{lyu2020towards} proposed a local DP based framework that privatizes each word's representation vector and sequentially replaces sensitive words in the text with semantically similar words. The key idea is to map each real value of the embedding vector into a binary vector with a fixed size and then privatize the vector via a variant of the unary encoding mechanism. Plant \textit{et al.} \cite{plant2021cape} developed the context-aware private embeddings (CAPE) approach that extracts the representation of each token from the final representation layer of a pre-trained model, normalizes it with sequence, and adds Laplace noise. Zhou \textit{et al.} \cite{zhou2023textobfuscator} further introduced TextObfuscator that obscures word information while maintaining word functionality through random perturbations applied to clustered representations. Clusters are created by identifying prototypes for each word and promoting word representations to be close to their respective prototypes. After training, words of similar functionality are close to the same prototype.  Random perturbations are applied to these clustered representations to protect privacy. They also introduced techniques for identifying prototypes for both token-level and sentence-level tasks by leveraging semantic and task-related information. However, in scenarios where the architecture and parameters of the LLM are known, there is no need to project the perturbed input into text space. 

Feyisetan \textit{et al.} \cite{feyisetan2020privacy} developed a text perturbation approach  to achieve the metric LDP. The metric LDP inherits the idea of local DP to ensure that the outputs of any two adjacent inputs are indistinguishable to protect the original input from being inferred. Furthermore, the metric LDP also aims to preserve the utility of sanitized texts by assigning higher probabilities to words that are semantically closer to the original ones. The developed approach \cite{feyisetan2020privacy} applies the generalized planar Laplace mechanism \cite{wu2017bolt} to perturb each token embeddings and further post-processes them to sanitized text via nearest neighbor search. To improve the performance, Xu \textit{et al.} \cite{xu2020differentially} used the Mahalanobis norm to replace the Euclidean norm adopted in \cite{feyisetan2020privacy} to measure the semantic similarities between words. Carvalho \textit{et al.} \cite{carvalho2023tem} proposed an improved perturbation method via the truncated Gumbel noise. 
Yue \textit{et al.} \cite{yue2021} further proposed utility-optimized metric LDP based on the observation that different inputs have different sensitivity levels to achieve higher utility. The developed SANTEXT+ approach divides all the text into a sensitive token set and an insensitive token set and allocates different privacy budget to each set. After deriving token vectors, it samples new tokens via metric LDP when the original tokens are in the sensitive set. 

Injecting DP noise into representation vectors directly may significantly distort the semantics of the original text.  Several research works \cite{chen2023customized,tong2023privinfer} leveraged the exponential mechanism to privately  replace each word or token in the raw prompt with semantically similar alternatives. Chen \textit{et al.}  \cite{chen2023customized} developed a customized text sanitization mechanism (CusText) that assigns each input token a customized output set of a small size and adopts the exponential mechanism to sample the output for each input token. The designed scoring function for exponential mechanism takes into account the semantic similarities between tokens during sampling.  Tong \textit{et al.} \cite{tong2023privinfer} developed the InferDPT framework and proposed RANdom adjacency for Text perturbation (RANTEXT) that introduces random adjacency for token-level perturbation of uploaded prompt. To provide strong protection for the raw prompt, its perturbation module utilizes the exponential mechanism to sequentially replace each word or token in the raw prompt with semantically similar alternatives from the adjacency list.  It also adopts the Laplace mechanism to dynamically determine the size of the adjacency list for each token. 

One limitation with the word-level protection is its lack of contextualization. The text generated by the perturbed prompt is often partially inconsistent and semantically incoherent with the raw prompt. This is because word-wise perturbation is conducted independently. Research has been proposed to address this challenge. For example, the InferDPT framework developed in \cite{tong2023privinfer} incorporates an extraction module that employs a local language model to extract text from the perturbed generation results so that the final reconstructed output is coherent, consistent, and aligns with the raw prompt. Note that the local language model is smaller than the remote LLMs and does not pose any privacy leakage. 

\subsubsection{Sentence Level Perturbation}
The word-level privacy, i.e., having each word indistinguishable with similar words, may not hide higher level concepts in the prompt. There have been a few research works studying sentence-level private embeddings. Sentence embeddings can be aggregated from token embeddings, for example, via mean pooling, or extracted from language models like BERT \cite{devlin2018bert}.  
Du \textit{et al.} \cite{du2023sanitizing} initiated a study of sanitizing sentence embeddings to achieve the metric LDP. They proposed two instantiations from the Euclidean and angular distances.  The former directly draws replacements from a distribution defined on a sphere and utilizes the Purkayastha mechanism \cite{weggenmann2021differential} based on the angular distance. The latter is to post-process the output of the noisy embedding by the Euclidean-distance based planar Laplace mechanism \cite{wu2017bolt}.   

Du \textit{et al.} \cite{du2023} considered sentence-level privacy for private fine-tuning. The proposed DP-Forward DP-Forward directly perturbs embedding matrices in the forward pass of PLMs and ensures the standard LDP for test sequences. They considered transformer-based LLMs which contain two categories of layers: embedding layers and task layers. The authors proposed an analytic matrix Gaussian mechanism (aMGM) to draw a non-iid DP noise from a matrix Gaussian distribution. DP-forward adds aMGM noise to embeddings output by layers preceding the task layers.  During inference, a user with an unlabeled sequence accesses the embedding layers to derive embedding matrices and perturbs them with aMGM noise. The user then sends the noisy embeddings to the LLM service provider. To make predictions, the LLM service provider runs the task layer functions on the noisy embeddings. With known LLM architectures and parameters, the perturbed embeddings can be directly utilized in white-box scenarios. Although their original purpose is to fine-tune models, these methods can also be applied for inference. 

\begin{figure*}[htp]
    \centering
    \includegraphics[width=18cm]{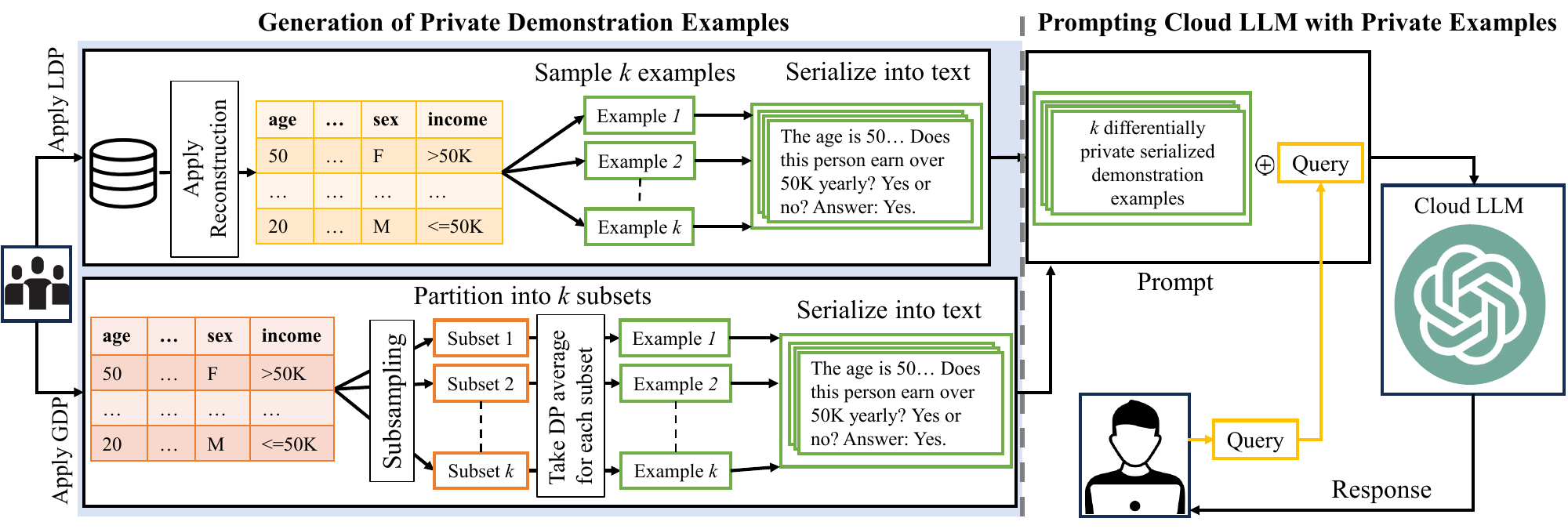}
    \caption{Privacy-preserving demonstration example generation with LDP-TabICL and GDP-TabICL approaches. For LDP-TabICL (top-left), users perturb their data with the randomized response LDP mechanism before being collected. The collected data is then reconstructed to recover the original data distribution. $k$ samples are selected and serialized into text. Meanwhile, for GDP-TabICL (bottom-left), user data is collected in clear. Then the collected data is partitioned into $k$ disjoint subsets. GDP averages for each attribute in each subset are generated. The generated noisy attributes are then serialized into text. During ICL an LLM is prompted with $k$ demonstration examples selected from the serialized text and a query from a user. LLM's response is generated and sent to the user.}
    \label{fig:dptabicl}
\end{figure*}
\subsubsection{Document Level Perturbation}
Instead of focusing only on word-level and sentence-level privacy, another work considers document-level privacy. 
To protect the privacy of the entire prompt during the inference process, Utpala \textit{et al.} \cite{utpala2023locally} devised DP-Prompt to attain document-level DP. DP-Prompt takes a private document and generates a paraphrased version using zero-shot prompting on a local pre-trained language model. 
The process of sequentially generating text from the language model is regarded as a problem of selecting tokens at each step. To make the generation step differentially private, DP-Prompt replaces the sequence with a differentially private version of the selection process. The resulting paraphrased document is then released as a sanitized document to prompt cloud LLMs. Specifically, given a private document alongside a designated prompt template instructing the language model, DP-Prompt generates text in a differentially private manner, producing a paraphrased version of the private document. The exponential mechanism is utilized for token selection during the sequential text generation process.

\subsubsection{Demonstration Example Level Perturbation}

The capabilities of LLMs have been extended to include tabular data analysis, leveraging the principles of ICL and prompt-tuning \cite{hegselmann2023, lin2024}. In performing this task, the common practice involves first serializing the tabular data into text before using them to prompt the LLM. To protect the privacy of demonstration examples, Carey \textit{et al.} \cite{carey2024} investigated the application of DP mechanisms for private tabular ICL via data privatization prior to serialization and prompting. They introduced the LDP-TabICL framework that employs the randomized response (RR) LDP technique \cite{warner1965} to perturb attribute values of each sample at the each end user side. Generating private demonstration examples within the LDP-TabICL involves reconstructing the DP-protected data obtained from users, selecting $k$ samples from the reconstructed data, and serializing the $k$ samples into texts. Similarly, concatenating the serialized texts with a query to facilitate ICL. The top-left part of Figure \ref{fig:dptabicl} depicts the private demonstration example generation in LDP-TabICL.  

\subsection{Global DP Methods}

\begin{figure*}[htp]
    \centering
    \includegraphics[width=18cm]{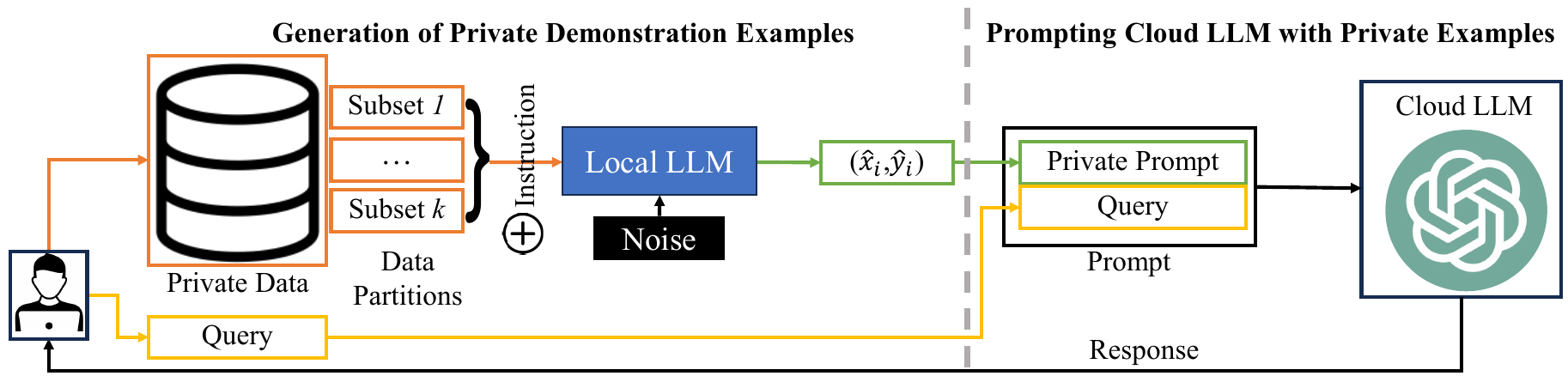}
    \caption{An illustration of privacy-preserving demonstration examples generated with a local LLM, then leveraging the examples with a query to perform ICL using a cloud LLM.}
    \label{fig:ppde}
\end{figure*}

In ICL, users often have a private local dataset  $D_{priv} = \{(x_i, y_i)\}_{i=1}^{n}$, from which a set of $k$ demonstration examples $D_k$ is chosen to be included in the prompt as shown in Equation \ref{eq:icl}. However, the simple inclusion of $D_k$ in the prompt certainly incurs the privacy disclosure. The cloud LLM is privacy-untrusted and may try to gain private information from the user's prompt. Global DP based algorithms can be adopted here to ensure that each individual sample in  $D_{priv}$ cannot be inferred with high confidence from the prompt sent to untrusted LLMs. One common solution is to generate the differentially private demonstration examples (denoted as $\hat{D}_k = \{(\hat{x}_i, \hat{y}_i)\}_{i=1}^{k}$) before employing them to perform ICL in the cloud LLM. Achieving global differential privacy here means the presence or absence of any example  $(x_i, y_i)$ in  $D_{priv}$ would not have significant impact on the produced $\hat{D}_k$ which will be used as demonstrations in ICL. Several methodologies have surfaced for generating differentially private demonstration examples: Sample and Aggregate based approach, PATE-based approach,  DP synthetic data generation approach, and soft prompt generation via DPSGD. 

\subsubsection{Sample and Aggregate based Approach}

Figure \ref{fig:ppde} presents an illustration of the concept. In broad terms, the process entails partitioning the private data into distinct subsets. These subsets are then submitted alongside instructions, guiding the local LLM to sequentially generate data resembling the private data token-by-token. At each token generation step, DP noise is introduced to the token probability. Consequently, the resultant generated data is DP private. This iterative process is repeated multiple times. Upon completion, the DP-protected examples are utilized to prompt the more powerful (yet untrusted) cloud LLM.

Tang \textit{et al.} \cite{tang2023} studied how to conduct in-context learning with LLMs on private datasets and focused on the privacy protection of demonstration examples used in the prompt. Their developed algorithm generates synthetic differentially private few-shot demonstrations from the original private dataset, and uses the generated samples as demonstrations in ICL during inference. The approach leverages the capabilities of local trusted LLMs in terms of generating a data sample similar to the ones in the original dataset. 
To generate a differentially private demonstration example for a given label $y$, the algorithm generates one token at a time from an empty list. At each token generation, disjointed subsets are extracted from the private training dataset $D_{priv}$. Each subset is appended with previously generated tokens and is fed into a local LLM to generate the next token. Next-token generation probabilities obtained from each subset are then privately aggregated. Both the Gaussian mechanism and report-noisy-max with exponential mechanism are adopted in the algorithm. Finally, the next token is produced and appended to previously generated tokens. This process is continued until the end of sequence token is produced. To reduce the effect of noise, the algorithm limits the vocabulary to the tokens present in top-K indices of the next-token probability coming from only the instruction without any private data.  
The generated private demonstration examples are leveraged to perform ICL on the LLM in the cloud (a less trusted environment). Note that the generated samples can be used for an infinite number of queries without incurring any additional privacy costs.

Similarly, to generate private demonstration examples locally, Hong \textit{et al.} \cite{hong2023} developed a framework called Differentially-Private Offset Prompt Tuning (DP-OPT). Given the private training dataset $D_{priv}$, DP-OPT uses a few samples as demonstrations to guide a local LLM  to generate private prompts. The prompt generation process is facilitated by a differentially private ensemble of in-context learning with disjoint private demonstration subsets. 
DP-OPT adopts the \textit{forward-backward} approach of Deep Language Network (DLN) \cite{sordoni2023}. This approach mimics the gradient-based optimization method that uses forward and backward passes to train prompts on a training dataset. Given the private training dataset $D_{priv}$, in the \textit{forward} pass of DP-OPT, the local LLM is prompted to predict the labels on a sample batch of training samples  $S \subset D_{priv}$ by submitting a forward template  with a task instruction $\pi$.  The output from the forward pass $\hat{y}_i$ is then used together with $S$ in a \textit{backward} template  to guide the local LLM. This backward pass attempts to regenerate the task instruction $\pi$. The regeneration of $\pi$ is performed one token at a time. During each token generation, token candidates are generated and aggregated with aggregates perturbed using differential noise. The best performing $\{(x_i,y_i,\hat{y}_i)\}$ are then privately selected and used for performing ICL in the cloud LLM.

\begin{figure}[htp!]
    \centering
    \includegraphics[width=\columnwidth]{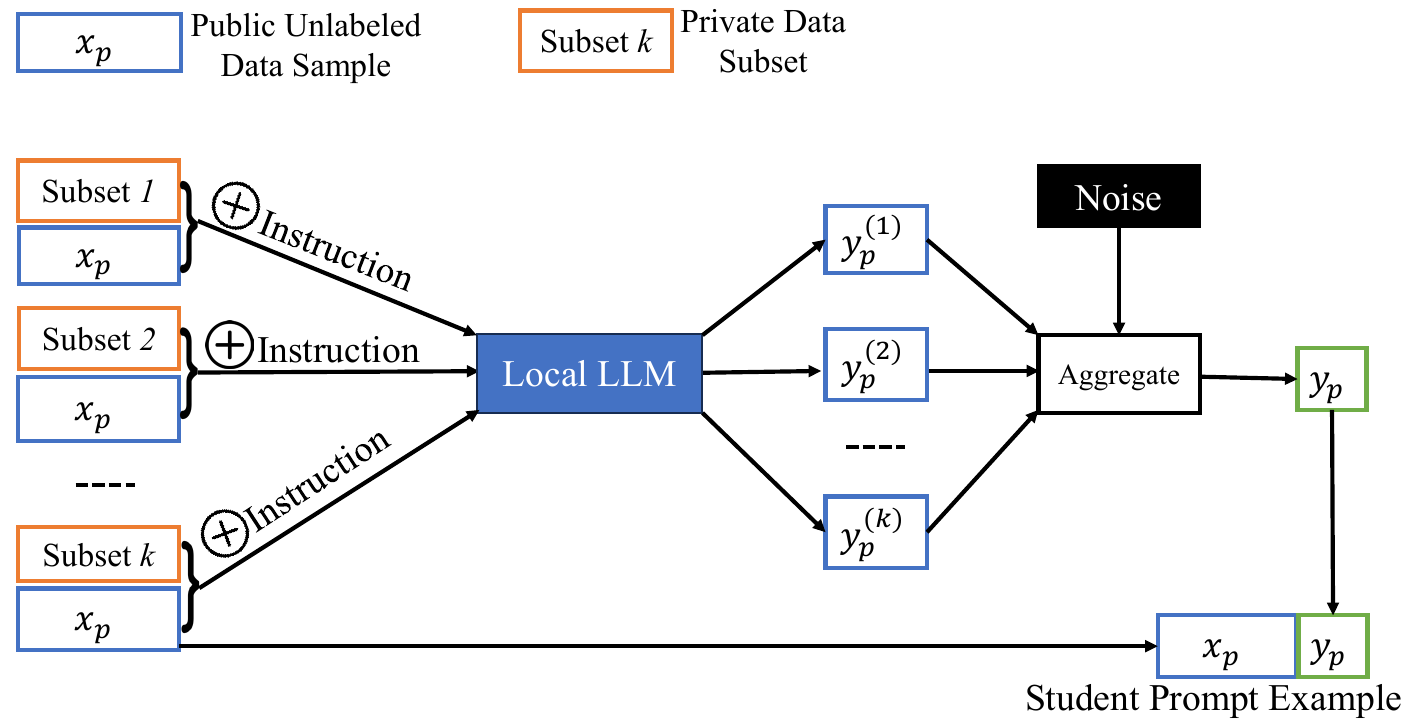}
    \caption{Private demonstration example (student prompt example) generation using PromptPATE \cite{duan2023flocks}. A public unlabeled data $x_p$ is labeled by prompting a local LLM with subsets of private data as prompt examples. A noisy aggregation is then performed on all prompt votes to generate a label $y_p$ for $x_p$. ($x_p, y_p$) can then be used to perform ICL publicly.}
    \label{fig:promptpate}
\end{figure}

\subsubsection{PATE-Based Approach} PATE \cite{papernot2018}  is a differential privacy learning model using a teacher-student framework. The student accessing unlabeled non-sensitive data distills knowledge from the aggregated predictions of multiple teachers. Each teacher model is trained on a partition of sensitive data. Calibrated noise is added to the aggregated predictions to meet DP requirements. 
Duan \textit{et al.} \cite{duan2023flocks} proposed PromptPATE, a privacy-preserving in-context learning framework for discrete prompts based on PATE \cite{papernot2018} mechanism.  It assumes the existence of a labeled private dataset with labeled examples like (``The book was great'', ``positive''), and an unlabeled public dataset with examples like (``I enjoyed this movie'',-). The private training dataset is divided into multiple subsets to create prompts that can be deployed with the LLM as teachers.  During the private knowledge transfer, for any input sequence from the unlabeled public dataset, each teacher votes for the most likely class. The consensus over the teachers' votes is determined via a noisy argmax over all teachers' vote counts. The added noise is sampled from a Gaussian distribution to satisfy the DP guarantees.  
The labeled public dataset is then leveraged as demonstration examples to perform ICL. The process is illustrated in Figure \ref{fig:promptpate}. 

Tian \textit{et al.} \cite{tian2022seqpate} developed the SeqPATE framework for text generation. Instead of sequential generation, SeqPATE generates pseudo-data using a pre-trained language model such that teachers only need to provide token-level supervision given the pseudo input. To address the large output space, SeqPATE aggregates teachers' outputs by interpolating their output distributions instead of voting for the final aggregate output. It also incorporates strategies to dynamically filter candidate words and only keep words  with high probabilities. Specifically, for the text generation task that aims to generate the remaining part of a sentence given its prefix,  SeqPATE assumes a private dataset (containing complete sentences) and a public dataset (containing input prefixes). SeqPATE trains each teacher model on one of the disjoint subsets of the private dataset and conducts student training and teacher inference on pseudo sentences generated by the large language models based on the public data. The student model is supervised by the private aggregation of teacher output distributions. 

Li \textit{et al.} \cite{li_2023_ICCV} proposed Prom-PATE that explores the benefits of visual prompting in generating image samples with DP noisy labels. Such generated samples can be employed to perform prompting in a DP manner. Specifically, Prom-PATE employs two steps to generate image samples with noisy labels: training re-teacher models and executing private aggregation. Training the re-teacher model involves training the soft prompts using a private dataset while maintaining the pre-trained visual model parameters unchanged. Several re-teacher models are trained using disjoint partitions of the private dataset. In the second step, Prom-PATE labels an unlabeled public image dataset by employing the PATE \cite{papernot2018} mechanism to perform DP aggregation on responses from re-teacher models. Re-teacher responses are generated with visual prompts. This step generates noisy labels for the samples in the unlabeled public dataset. Such noisy labeled samples can then be used to freely perform prompting. However, such studies are relatively rare and more work is required to determine their effectiveness.

\subsubsection{DP Synthetic Data Generation}\label{sec:synthetic_data}
One notable advantage of differentially private synthetic data generated from a private dataset is that the resulting text can be freely shared and utilized (including for performing private ICL) with minimal privacy concerns. Yue \textit{et al.} \cite{yue2022} and Flemings \textit{et al.} \cite{flemings2024} fine-tuned a pre-trained generative language model with DP using a private dataset, enabling the models to produce synthetic text with robust privacy protections. The studies utilized the DP-SGD \cite{abadi2016} framework during the fine-tuning process. DP synthetic data is generated by prompting the fine-tuned model with control codes. This DP synthetic data captures the general statistical characteristics of the private text and can be utilized more freely for various downstream tasks, with minimal privacy risks. For example, \cite{flemings2024} used the DP synthetic data alongside a teacher model output distribution to transfer knowledge from the teacher model to a student model.  
Kurakin \textit{et al.} \cite{kurakin2023} chose to perform parameter-efficient fine-tuning on the pre-trained LM using prompt tuning \cite{lester2021} and LoRA \cite{hu2021}. The authors still adopted the DP-SGD framework during the fine-tuning. In this case, DP synthetic data is generated by feeding prefix as input to the fine-tuned model. Carranza \textit{et al.} \cite{carranza2023} adopted the same approach to generate DP queries to train retrieval systems. Their work employed DP-Adafactor (Adafactor \cite{shazeer2018} that receives clipped and noised gradients as per DP-SGD \cite{abadi2016}) to fine-tune a pre-trained LM. The DP-tuned LM is then used to generate DP synthetic queries.

Xie \textit{et al.} \cite{xie2024} considered generating DP synthetic data with only API access to LLMs. They proposed an augmented Private Evolution (PE) framework referred to as AUG-PE. The concept behind AUG-PE involves initially sampling random instances from an LLM guided by instructions, followed by iterative enhancements through DP selections, focusing on those resembling the private dataset. Subsequently, the LLM is queried to generate additional samples resembling the selected ones. The entire concept can be divided into four steps, with steps 2-4 carried out iteratively. In step 1, an LLM is prompted to generate random samples. Subsequently, in step 2, each private sample votes for its nearest synthetic counterpart in the embedding space, followed by the addition of Gaussian noise to these votes. This produces a DP nearest neighbor histogram. With the help of this histogram, step 3 involves the selection of synthetic samples with noisy votes. Finally, in step 4, the LLM is prompted to generate new samples resembling the noisy selection in step 3.

Meehan \textit{et al.} \cite{meehan2022sentence} proposed DeepCandidate to achieve sentence-level differential privacy when releasing a document embedding.  The document embedding is sentence private if any single sentence in the document is removed or replaced we can still have a similar probability of producing the same embedding. Thus, the sentence-level privacy is defined from the global DP perspective, i.e., hiding the impact of any single sentence in a document. The privatized document embedding only stores limited information unique to any given sentence. DeepCandidate uses a sentence encoder to get sentence embeddings and adopts the exponential mechanism to sample from the candidate embeddings for each private embedding. 

Emerging alternative methodologies avoid the need for local (trusted) LLM or encoder support. Carey \textit{et al.} \cite{carey2024} embraced this approach for conducting tabular data analysis with a DP guarantee. To protect the privacy of demonstration examples, \cite{carey2024} developed the GDP-TabICL framework for private tabular ICL via data privatization prior to serialization and prompting.  GDP-TabICL relies on both Poisson subsampling for privacy amplification and the Laplace mechanism to craft differentially private aggregates that represent the underlying data distribution. GDP-TabICL segments the sampled data into $k$ disjoint subsets based on the required number of demonstration examples. DP statistics for each feature in each subset are subsequently computed. These statistics are used to generate synthetic examples and then serialized into text-based demonstration examples along with a test query to facilitate ICL. Figure \ref{fig:dptabicl} depicts the proposed approach.
Different from works \cite{duan2023flocks, tang2023, hong2023} that assume the use of local LLMs to produce differentially private prompts, DP-TabICL \cite{carey2024} uses reconstruction from perturbed statistics to generate DP demonstration examples and thus incurs much less computational cost. 

\subsubsection{Soft Prompt via DPSGD} 
\begin{figure}[htp]
    \centering
    \includegraphics[width=8cm]{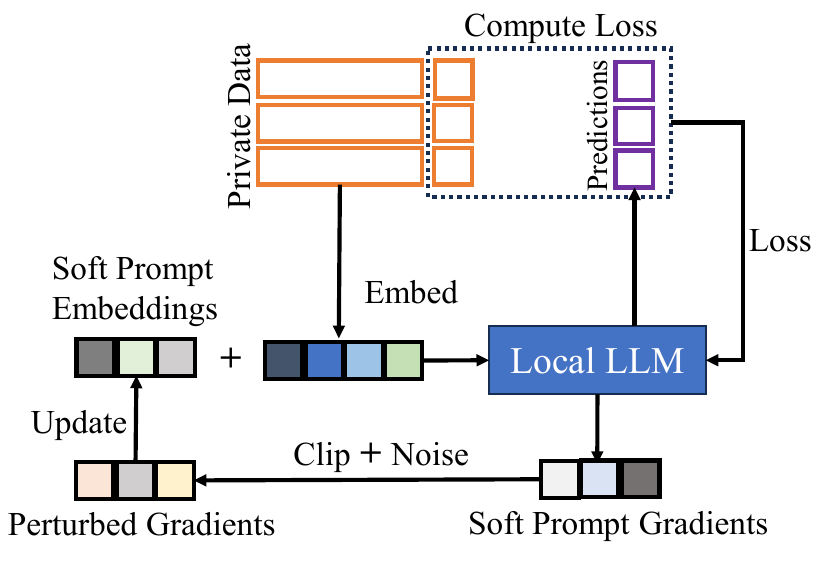}
    \caption{An illustration of private soft prompt generation using PromptDPSGD \cite{duan2023flocks}. A soft prompt is prepended to the private data embeddings. This combination is forwarded to an LLM. The LLM makes predictions $\hat{y}$ and computes loss accordingly. Next, LLM uses the loss to compute the gradients associated with the soft prompt. These gradients are then clipped, and noise is added to them before being used to update the soft prompt privately.}
    \label{fig:promptdpsgd}
\end{figure}

While the above studies focus on discrete prompts that require only black-box access to the LLMs, approaches that privately learn soft prompts are also  important. Soft prompts are additional task specific embeddings that can be prepended to the original input embeddings before passing them through the LLMs \cite{duan2023flocks}. The gradient descent method can be employed to train these embeddings using private data. For private training, DP noise is added to the gradients associated with these embeddings. Duan \textit{et al.} \cite{duan2023flocks}, developed PromptDPSGP that leverages the DP-SGD algorithm \cite{abadi2016} to learn soft prompts that are prepended to an LLM's input with a differential privacy guarantee. An illustration of the process is shown in Figure \ref{fig:promptdpsgd}. However, soft prompts require white-box access to LLMs, so this may not be possible all the time.

\subsection{Other Scenarios}

\subsubsection{Demonstration Examples at LLMs}
\begin{figure*}[htp]
    \centering
    \includegraphics[width=18cm]{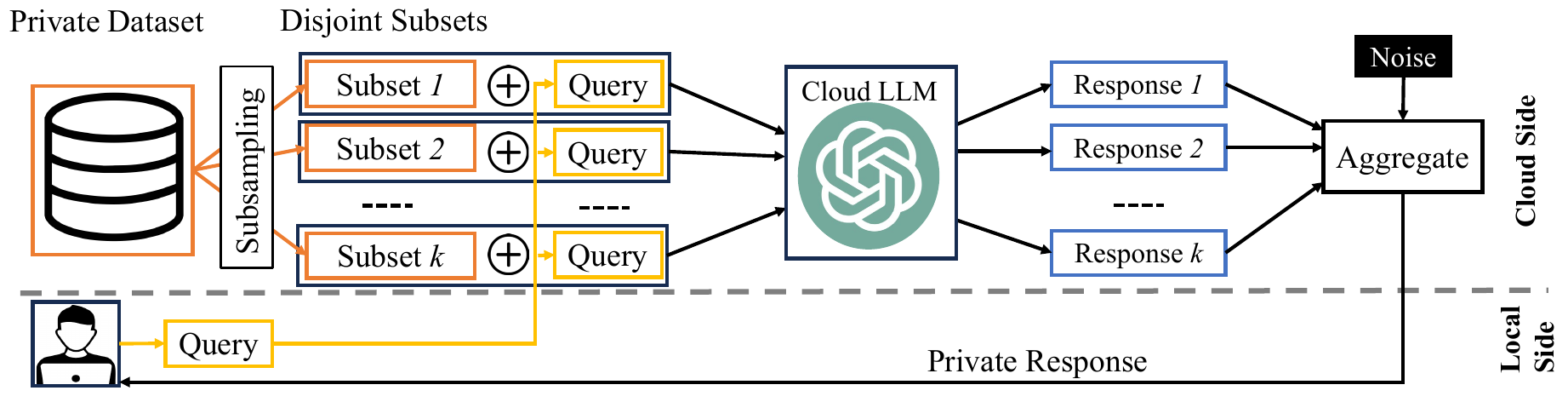}
    \caption{Differentially private ICL (DP-ICL) with a private response. Disjoint subsets of a private dataset are each used as prompt examples to answer a query. Responses to each prompt are privately aggregated to generate the final response to the query.}
    \label{fig:dpicl}
\end{figure*}

An alternative point of leakage is through model outputs. An adversary can infer information about the input data from model outputs. A solution can be achieved through a noisy consensus among an ensemble of an LLM's responses.
Wu \textit{et al.} \cite{wu2023privacy} proposed a differentially private in-context learning (DP-ICL) paradigm where the sensitive dataset used for demonstrations is stored in the LLM site. The server partitions the sensitive dataset into disjoint subsets, each comprising a collection of demonstration examples. The server generates these demonstration-query pairs and calls LLM to produce corresponding outputs. These outputs are aggregated through a differentially private mechanism before being returned to the user. Figure \ref{fig:dpicl} depicts this framework. 
For text classification, the authors \cite{wu2023privacy}  adopted Report-Noisy-Max with Gaussian noise to privately release the class that receives the majority vote. For language generation, to deal with the challenge arising from the huge output sentence space, the authors proposed 1) termed Embedding Space Aggregation (ESA) which projects the output sentences into a semantic embedding space and then privatizes these aggregated embeddings; and 2) Keyword Space Aggregation (KSA) which identifies frequently occurring keywords in the output and then privately selects them via propose-test-release \cite{dwork2009differential} or the joint exponential mechanism \cite{gillenwater2022joint}. 
Different from DP synthetic data generation algorithms, DP-ICL does not permit an infinite number of queries as each prompt query incurs privacy consumption.

\subsubsection{Data Augmentation using External Datastores}
\label{sec:database-augmentation}

For PLM users with insufficient local data, leveraging external datastores can improve  ICL performance. The final generated prompt may contains sensitive information from both the local data and external datastores. Retrieval-based techniques \cite{khandelwal2019, borgeaud2022, izacard2022, min2022}, which have been developed to combine LLMs' output with retrieved texts from datastores, can also be applied here to help users prepare augmented prompts. However, this process incurs potential leakage of sensitive information of the external datastores.  This raises the obvious question: \textit{Can we introduce external knowledge in the prompt without compromising the privacy of the external datastores?} We briefly review sanitization and information flow control methods that can be used to address this question. 

As stated in section \ref{sec:sanitization}, sanitization aims to create a sanitized version that would keep structural properties but remove sensitive information from the original data. A common approach used in machine learning involves masking the sensitive data attributes \cite{biesner2022, nakamura2020, lukas2023}. 
Huang \textit{et al.} \cite{huang2023} proposed a sanitization method for replacing each privacy-sensitive phrase in the datastore  with: (i) $<|\text{endoftext}|>$, 
(ii) dummy text (e.g., replacing a telephone contact with ``012-345-6789"), or (iii) public data (e.g., replacing a telephone contact with a publicly known telephone number). Information flow control (IFC) is a privacy-by-design model.  Wutschitz \textit{et al.} \cite{wutschitz2023} developed an IFC-based framework that takes into consideration metadata such as access control policies in designing machine learning systems. Private external data retrieved during the prompt augmentation is subject to satisfying security requirements (e.g., access policies). This ensures that the information presented to the users is what they are authorized to see. For example, when selecting demonstration examples from an external store, the model ensures the retrieved samples satisfy the security requirements. Under the scenarios where separate public corpora and private corpora exist, Xiong \textit{et al.} \cite{xiong2020} and Arora \textit{et al.} \cite{arora2023} developed multi-hop retrieval-based models to protect privacy.  The retrieval process is performed iteratively, limiting information flow and careful ordering of the retrieval from both public and private corpora.  

\subsubsection{Client Data Protection via FL}
Utilizing pre-trained models through prompting has also been expanded to include visual foundational models \cite{jia2022visual, zhou2022learning}. For example, Jia \textit{et al.} \cite{jia2022visual} explored soft prompts by prepending them on the input tokens of pre-trained visual transformers. In this manner, the soft prompts are optimized to capture task-specific details, enabling them to be used to prompt the pre-trained model to generate suitable responses tailored to the task at hand. Consequently, there is a rising interest in studies that aim to tackle the privacy challenges associated with visual prompting (VP) \cite{yang2023efficient, su2024federated, guo2023promptfl}.
The majority of works in this area consider the adoption of federated learning (FL) approach. FL enables clients to learn a joint model without sharing their private data \cite{mcmahan2017communication}. 

Guo \textit{et al.} \cite{guo2023promptfl} proposed the PromptFL framework which is built upon the federated learning mechanism. The PromptFL framework operates on the basis that each client possesses a CLIP foundation model \cite{pmlr-v139-radford21a}. Within PromptFL, clients keep their data locally and train shared soft prompts collaboratively by communicating gradients rather than the data. This approach facilitates the development of robust prompts while safeguarding the privacy of individual client data. Given the diversity in clients' data, personalized federated learning methods \cite{chen2021bridging, shamsian2021personalized} have also emerged. These approaches enable individual clients to train personalized models tailored to their specific datasets.
Su \textit{et al.} \cite{su2024federated} and Yang \textit{et al.} \cite{yang2023efficient} proposed federated adaptive prompt tuning algorithm (FedAPT) and personalized FL for client-specific prompt generation (pFedPG) frameworks, respectively. These models adopt the personalized FL approach. Specifically, in FedAPT, each client trains an adaptive network and prompts using their private data along with a random key assigned by the server. A global adaptive network and global prompts are computed on the server using adaptive networks and prompts received from the clients. These global parameters are then utilized to generate personalized prompts for each client. Meanwhile, pFedPG learns a personalized prompt generator at the server, which is used to generate client-specific prompts. The framework consists of two phases: global personalized prompt generation and local personalized prompt adaptation. During the personalized prompt adaptation phase, each client trains client-specific prompts. On the other hand, personalized prompt generation is achieved by learning to derive personalized prompts for each client through the exploitation of optimization directions among clients.   
The principles underlying these frameworks offer valuable insights for achieving client-level privacy in NLP contexts, warranting further exploration in the domain.

\section{Resources}\label{sec:resources}

\subsection{Datasets}
\label{sec:dataset}

\begin{table*}
\begin{center}
    \begin{tabular}{|c|ccccc|}
    \hline

    \hline
        Type & NLP Task Category &Dataset & Ref & No. of Samples & Citing Publications \\
    \hline

    \hline
        \multirow{24}{*}{Text Datasets} & \multirow{12}{*}{Classification} & AGNews & \cite{zhang2015} & 496835 & \cite{duan2023, tang2023, panda2023, duan2023flocks}\\
        & & DBPedia & \cite{dbpedia} & 342781& \cite{tang2023, duan2023flocks}\\
        & & TREC & \cite{li-roth-2002-learning}& 6000&\cite{duan2023, tang2023, panda2023, duan2023flocks, hong2023}\\
        & & SST-2 & \cite{socher2013} & 11855&\cite{duan2023, panda2023, duan2023flocks, hong2023}\\
        & & MPQA & \cite{wiebe2005} & 10657&\cite{hong2023}\\
        & & Disaster & \cite{nlp-getting-started} & 10746 &\cite{hong2023}\\
        & & CB & \cite{cb} & 556&\cite{duan2023}\\
        & & QNLI & \cite{wang2018} & 110400&\cite{duan2023flocks}\\
        & & QQP & \cite{wang2018}& 755000&\cite{duan2023flocks}\\
        & & MNLI & \cite{wang2018} & 413000&\cite{duan2023flocks}\\
        & & BBC & \cite{learn-ai-bbc} & 2225&\cite{chen2023}\\

        \cline{2-6}
        & \multirow{7}{*}{Information Extraction} & MIT-G & \cite{liu2012} &3053& \cite{tang2023}\\
        & & MIT-D & \cite{liu2012} &1661& \cite{tang2023} \\
        & & Elsevier & \cite{kershaw2020} &40091&\cite{wutschitz2023}\\
        & & Arxiv & \cite{arxiv}&About 1.7 Million&\cite{wutschitz2023}\\
        & & ACE2005 & \cite{ace2005} &About 1800 &\cite{kan2023}\\
        & & Enron Emails & \cite{klimt2004} &About 500000&\cite{huang2023}\\
        & & WikiText & \cite{merity2016} &Over 100 Million&\cite{huang2023}\\

        \cline{2-6}
        & Creative Writing & WritingPrompts & \cite{fan2017} &10700 &\cite{zhang2023latticegen}\\

        \cline{2-6}
        & \multirow{2}{*}{Question Answering} & ConcurrentQA& \cite{arora2023} &18439& \cite{arora2023} \\
        & & DocVQA & \cite{tito2023} &46000& \cite{panda2023}\\

        \cline{2-6}
        &Summarization & SAMSum & \cite{gliwa2019} &16369& \cite{panda2023} \\

        \cline{2-6}
        &Recommendation & Amazon (beauty) & \cite{amazon} &Over 2 Million & \cite{lin2024} \\
    \hline
        \multirow{8}{*}{Tabular Datasets} & \multirow{8}{*}{Tabular Data Analysis} & Adult & \cite{misc_adult} &48842& \cite{lin2024, carey2024}\\
        & & Bank & \cite{bank} &45211&\cite{carey2024}\\
        & & Blood &\cite{blood} &748&\cite{carey2024}\\
        & & Calhousing & \cite{calhousing} &20640&\cite{carey2024}\\
        & & Car &\cite{kadra2021well} &1728&\cite{carey2024}\\
        & & Diabetes & \cite{diabetes} &768&\cite{carey2024}\\
        & & Heart & \cite{heart} &918&\cite{carey2024}\\
        & & Jungle & \cite{van2014endgame} &44819&\cite{carey2024}\\
    \hline
        \multirow{5}{*}{Image Datasets} & \multirow{5}{*}{Image Data Analysis} & Office-Caltech10 & \cite{griffin2007caltech, saenko2010adapting} &2533 & \cite{yang2023efficient, su2024federated}\\
        & & DomainNet & \cite{peng2019moment} &0.6 Million & \cite{yang2023efficient, su2024federated}\\
        & & Dermoscopic-FL & \cite{chen2021personalized} &10490 & \cite{yang2023efficient}\\
        & & CIFAR-10 & \cite{krizhevsky2009learning} &60000 & \cite{yang2023efficient, li_2023_ICCV}\\
        & & CIFAR-100 & \cite{krizhevsky2009learning} &60000 & \cite{yang2023efficient, li_2023_ICCV}\\
        & & Blood-MNIST & \cite{yang2023medmnist} &17092 & \cite{li_2023_ICCV}\\
    \hline

    \hline
    \end{tabular}
    \caption[Dataset table]{Commonly used datasets for evaluating privacy-preserving prompting frameworks\label{tab:datasets}}
\end{center}
\end{table*}

In addition to developing privacy-preserving frameworks, it is crucial to identify high-quality datasets for various tasks to evaluate the performance of these frameworks. This section provides an overview of widely used datasets for assessing these frameworks. Table \ref{tab:datasets} outlines the datasets, categorized according to data types and tasks.

\subsubsection{Text Datasets}
Here, we focus on commonly used text datasets to evaluate privacy-preserving prompting frameworks. We organize them according to tasks, as follows:

\textit{Classification:}
For text classification tasks, \cite{duan2023, tang2023, panda2023, duan2023flocks, hong2023}, and \cite{chen2023} use at least one of the following datasets to evaluate their privacy-preserving frameworks.

\begin{itemize}
    \item \textbf{AGNews} \cite{zhang2015} dataset consists of news articles from the web with 496,835 samples and it has four classes (\textit{world, sports, business} and \textit{sci/tech}).
    
    \item \textbf{DBPedia} \cite{dbpedia} dataset contains a structured extraction of Wikipedia information. It consists of fourteen classes (\textit{company, school, artist, athlete, politician, transportation, building, nature, village, animal, plant, album, film} and \textit{book}) and is mainly used for topic classification. 
    
    \item \textbf{TREC} \cite{li-roth-2002-learning} dataset is a question classification dataset with six classes (\textit{description, entity, expression, human, location} and \textit{number}). The train set consists of 5500 samples, while the test set has 500 samples. 
    
    \item \textbf{SST-2} \cite{socher2013} dataset consists of single-sentence movie reviews and their annotations for sentiments. It contains 11,855 samples with 215,154 unique phrases. It has two classes (\textit{positive} and \textit{negative}). 
    
    \item \textbf{MPQA} \cite{wiebe2005} dataset contains an annotation of 10,657 sentences from multiple news sources for opinions and other private details. The dataset is mainly used for sentiment classification. It has two classes (\textit{positive} and \textit{negative}).
    
    \item \textbf{Disaster} \cite{nlp-getting-started} dataset is used to classify if a text refers to a disaster event or not. It consists of two classes with a 7,503 sample train set and a 3,243 sample test set.
    
    \item \textbf{CB} \cite{cb} dataset is mainly used for sentiment analysis. It has three classes (\textit{contradiction, entailment} and \textit{neutral}). The training set has 250 samples, the validation set has 56 samples and the test set has 250 samples. 
    
    \item \textbf{QNLI} \cite{wang2018} dataset is a natural language inference (NLI) dataset consisting of a question-paragraph pair. The paragraphs comprise sentences drawn from Wikipedia and contain the answers to the questions. The training set contains 105,000 samples, while the test set contains 5,400 samples. 
    
    \item \textbf{MNLI} \cite{wang2018} dataset is also an NLI dataset that contains an annotated crowd-sourced collection of sentences. It is mainly used for sentiment analysis and it contains three classes (\textit{entailment, neutral} and \textit{contradiction}). The training set contains 393,000 samples while the test set contains 20,000 samples. 
    
    \item \textbf{QQP} \cite{wang2018} dataset consists of question pairs. Each question pair is annotated with a binary value to indicate if one question is a paraphrased version of the other. The training set contains 364,000 samples while the test set contains 391,000 samples. 
    
    \item \textbf{BBC} \cite{learn-ai-bbc} dataset consists of 2,225 articles each labeled with one of the following classes: \textit{business, entertainment, politics, sport} and \textit{tech}). The training set consists of 1,490 samples while the test set has 735 samples. 
\end{itemize}

\textit{Information Extraction:}
Information extraction entails extracting structured information suitable for storage and analysis from unstructured texts. For the information extraction task, \cite{huang2023} uses the Enron Emails and WikiText datasets, \cite{tang2023} uses MIT Movies trivia10k13  dataset,  \cite{kan2023} uses ACE2005 dataset, and \cite{wutschitz2023} uses the Elsevier and Arxiv datasets.

\begin{itemize}
    \item \textbf{Enron Emails} \cite{klimt2004} dataset contains at least 500,000 emails from the 150 employees of the Enron Corporation.
    
    \item \textbf{WikiText} \cite{merity2016} dataset contains over 100 million tokens of verified articles from Wikipedia.  
    
    \item \textbf{MIT Movies trivia10k13} \cite{liu2012} dataset comprises movie reviews with slots to be filled. The slots are filled with either the movie genre (MIT-G) or the movie's director name (MIT-D). MIT-G contains 2,953 samples as a training set and 100 samples as a test set. MIT-D contains 1,561 samples as a training set and 100 samples as a test set.  
    
    \item \textbf{ACE2005} \cite{ace2005} dataset is a multilingual dataset contains approximately 1,800 files of mixed genre text in English, Arabic, and Chinese training data for automatic content extraction tasks. It contains data of different types annotated for entities, relations, and events.  
    
    \item \textbf{Elsevier} \cite{kershaw2020} dataset contains 40,091 journal articles from multiple scientific disciplines. Each article contains the article abstract, the article body, and the article's author information. 
    
    \item \textbf{Arxiv} \cite{arxiv} dataset consists of PDF files of over 1.7 million open-access scientific articles. Similar to the Elsevier dataset, each Arxiv article contains an abstract, the main text body, and author information.
\end{itemize}

\textit{Other NLP Tasks:}
We cover the other NLP tasks as follows: for question-answering tasks, \cite{panda2023} uses the DocVQA dataset, and \cite{arora2023} uses their own ConcurrentQA dataset. For the creative writing task, which aims to dynamically select a sequence of events, actions, or words that collectively form a cohesive story, \cite{zhang2023latticegen} uses the WritingPrompts dataset. For the summarization task, which condenses a large piece of test, \cite{panda2023} uses the SAMSum dataset. And, for the recommendation task, which provides users with suggestions for actions and items based on their preferences and past interactions, \cite{lin2024} uses the Amazon (beauty) dataset. 

\begin{itemize}
    \item \textbf{DocVQA} \cite{tito2023} dataset consists of 46,000 questions posed by over 48,000 scanned images of industry documents. In \cite{panda2023}, the questions are answered using the extracted OCR tokens from the document images.
    
    \item \textbf{ConcurrentQA} \cite{arora2023} dataset contains questions traversing Wikipedia documents and Enron employee emails \cite{klimt2004}. The dataset contains 15,239 samples as a training set, 1,600 as a validation set, and 1,600 as a test set. 

    \item \textbf{WritingPrompts} \cite{fan2017} dataset consists of stories and their high-level descriptions formulated as prompts. It contains 10,000 samples as a training set and 700 samples as a test set.

    \item \textbf{SAMSum} \cite{gliwa2019} dataset consists of conversations between two individuals with some texts exchanged being private. It contains 16,369 conversations with annotated summaries. The training set consists of 14,732 samples, the validation set has 818 samples and the test set has 819 samples.

    \item \textbf{Amazon (beauty)} \cite{amazon} dataset contains reviews from users for Amazon beauty products. It has ratings ranging from 1-5 depending on customer satisfaction.
\end{itemize}

\subsubsection{Tabular Datasets}
For tabular data analysis, \cite{lin2024} uses the Adult dataset and \cite{carey2024} uses the following datasets. 

\begin{itemize}
    \item \textbf{Adult} \cite{misc_adult} dataset aims to predict if an individual's annual income exceeds $\$50,000$. The dataset has 48842 samples with 12 features.

    \item \textbf{Bank} \cite{bank} dataset is used to predict whether a bank client will subscribe to a term deposit. The dataset has 45211 samples with 16 features. 

    \item \textbf{Blood} \cite{blood} dataset is utilized for predicting whether a person will donate blood. It comprises 748 samples with 4 features.

    \item \textbf{Calhousing} \cite{calhousing} dataset predicts if a given housing block value is above the median based on location. The dataset has 20,640 samples with 8 features. 

    \item \textbf{Car} \cite{kadra2021well} dataset evaluates the state of each car. The dataset contains 1728 samples with 6 features.

    \item \textbf{Diabetes} \cite{diabetes} dataset is used to predict if a patient has diabetes or not. The dataset comprises 768 samples with 8 features.

    \item \textbf{Heart} \cite{heart} dataset is used to predict if a patient has heart disease based on coronary angiography. The dataset has 918 samples with 11 features.

    \item \textbf{Jungle} \cite{van2014endgame} dataset's goal is to predict if the white player wins a two-piece endgame of jungle chess. The dataset has 44819 samples with 6 features.

\end{itemize}

\subsubsection{Image Datasets}
To perform image data analysis as a downstream task, \cite{yang2023efficient, su2024federated, li_2023_ICCV} use at least one of the following image datasets.

 \begin{itemize}
     \item \textbf{Office-Caltech10} \cite{griffin2007caltech, saenko2010adapting} dataset comprises four data domains: Amazon, DSLR, Webcam, and Caltech, with each domain having 10 classes. The dataset has 2533 samples in total. 

     \item \textbf{DomainNet} \cite{peng2019moment} dataset encompasses six domains: Clipart, Infograph, Painting, Quickdraw, Real, and Sketch. It consists of 0.6 million samples with 345 classes distributed across these six domains.

     \item \textbf{Dermoscopic-FL} \cite{chen2021personalized} dataset comprises images from four data sites collected from HAM10K \cite{tschandl2018ham10000} and MSK \cite{codella2018skin}. Each site contains three types of skin lesions, resulting in a total of 10,490 samples. 

     \item \textbf{CIFAR-10 \& CIFAR-100} \cite{krizhevsky2009learning} datasets. The CIFAR-10 dataset contains 60000 images with ten classes in total. Each class has 6000 images. While the CIFAR-100 dataset has 60000 images with 100 classes. Each class has 600 images.

     \item \textbf{Blood-MNIST} \cite{yang2023medmnist} dataset comprises images of blood cells from uninfected patients, totaling 17,092 samples with eight classes.
 \end{itemize}

\subsection{Software Tools}
Strides have also been made to create software tools designed to offer privacy protection during interactions with LLMs through prompting. These tools are designed to be integrated with LLM applications, ensuring the privacy of prompts, demonstration examples, and LLM responses.

\subsubsection{Proprietary Software Tools} We list a few proprietary software tools below.

\begin{itemize}
    \item \textbf{Anonos Prompt Protector} \cite{anonos} prevents sensitive prompt data from leaking in LLMs by replacing sensitive attributes with dummies such as NAME\_1, AGE\_1, etc. These dummy attributes get restored to their original values in LLM responses. 
    
    \item \textbf{Prompt Security} \cite{prompt-security} monitors exchanges between users and LLMs for sensitive information. Once sensitive information is detected, it either sanitizes the sensitive attribute or blocks the information from being forwarded to the intended recipient. The tool also offers other services such as protection against prompt injection and jailbreak attacks. 
    
    \item \textbf{WhyLabs} \cite{whylabs} similarly monitors exchanges between users and LLMs. It evaluates prompts for prompt injection attacks and blocks LLM responses that contain PII. This way, the tool is able to present privacy-violating responses. However, it simply blocks prompts that contain malicious content.  
    
    \item \textbf{CalypsoAI Moderator} \cite{calypsoai} conducts work similar to that of WhyLabs. It assesses prompts and prevents them from being executed if they contain information that could lead to a privacy-violating response from the LLM. 
    
    \item \textbf{Lakera Guard} \cite{lakeraguard} sanitizes input data by replacing PII values with entity types such as EMAIL\_ADDRESS, CREDIT\_CARD, etc. It also provides additional services to mitigate prompt injection attacks and prevent the generation of harmful content. These targets are achieved by serving as an intermediary between users and LLMs. 
\end{itemize}

\subsubsection{Open Source Software Tools}
We now shift our attention to open-source software tools that aim to provide privacy-preserving prompting services. 
\begin{itemize}
    \item \textbf{LLM Guard} \cite{llmguard} developed by Protect AI is a tool that anonymizes input data to prevent PII leakage and de-anonymizes the response returned from the LLM to restore the sanitized attributes. It also provides capabilities for detecting harmful language and defending against prompt injection attacks, ensuring the safety and security of your interactions with LLMs.

    \item \textbf{Guardrail AI} \cite{guardrails} conducts an analysis of inputs to LLMs as well as their responses. Through this process, it detects, quantifies, and mitigates specific types of risks, thereby preventing the exposure of sensitive information like PII. 
\end{itemize}

\section{Limitations and Future Prospects}\label{sec:future_prospects}

\subsection{Limitations}\label{sec:lim}
Despite the strides made in tackling privacy challenges in prompting, current frameworks still exhibit weaknesses such as computational inefficiencies, semantic inadequacies, privacy and trustworthiness challenges. Here, we outline these weaknesses.

\subsubsection{Computational Inefficiency} 
Many of the frameworks discussed exhibit computational inefficiency. We elucidate on this as follows: 
(i) Sanitization-based frameworks. To identify and sanitize sensitive attributes in users' texts, the sanitization-based frameworks \cite{kan2023, chen2023, zhang2024cogenesis} rely on local LLMs. Running LLMs locally requires enormous amounts of computational power. An everyday user may not afford the luxury of having computational devices capable of running such models locally. 
(ii) Ensemble-only framework \cite{duan2023}. It requires multiple subsets of private data to be sent to LLMs along with the same query. This can pose both communication and computation challenges and increase inference time. 
(iii) Obfuscation/lattice-based frameworks. These frameworks attempt to conceal both prompts and responses from adversaries, and doing so comes with computation challenges. For example, in the lattice-based LatticeGen \cite{zhang2023latticegen} framework, each unique lattice configuration necessitates a distinct lattice-finetuned LLM. Fine-tuning for each lattice configuration is computationally demanding. Additionally, the lattice method is reliant on the exchange of tokens between the user and the server, the more tokens are exchanged, the more computation and communication resources are consumed. This gets worse for longer text sequence tasks such as creative writing. The obfuscation-based IOI framework \cite{yao2024} combines the target instance with obfuscators to ensure that the instance is never directly exposed to the LLM. To ensure robust privacy protection and stable task performance, strategies such as balancing and randomization involve emitting additional requests, leading to multiple inferences for each input instance. The additional inferences come with additional computational costs. 
(iv) Encryption-based frameworks. As is the case with obfuscation/lattice-based frameworks, the encryption-based frameworks protect both the prompts and the responses, and as a result, the same challenge extends to cryptographic encryption methods \cite{hou2023ciphergpt, hao2022iron, chen2022thex}. These frameworks perform collaborative computations during inference. The computations for the various non-linear functions of LLMs during inference require data exchange among multiple parties and entail other resource-intensive cryptographic computations. 
(v) Global DP based frameworks. The GDP based frameworks \cite{tang2023, hong2023, duan2023flocks} equally rely on local PLMs to facilitate DP noise addition. However, as with sanitization frameworks, the reliance on local PLMs necessitates computationally powerful devices, which may pose a barrier for everyday users.  

\subsubsection{Semantic Inadequacy}
A good number of the frameworks generate semantically inaccurate texts. We expound on this phenomenon by categorizing them according to their mechanisms as follows: (i) Obfuscation/lattice-based frameworks. These frameworks perturb their inputs and can lead to poor quality of the generated text at the LLM. For instance, the lattice-based framework LatticeGen \cite{zhang2023latticegen} indiscriminately adds noise at each token generation. This noise grows with the generated text length, thus misguiding the LLM into generating semantically inaccurate text. In addition, the obfuscation-based framework IOI \cite{yao2024} is not tailored for text generation and it results in poor text quality when used for text generation purposes.
(ii) Encryption-based frameworks. Similarly, with the emoji-based encryption framework EmojiCrypt \cite{lin2024}, the generated emojis might be misleading/misinterpreted. Emoji space is limited in comparison to the text vocabulary space. This may lead to an emoji being used to represent multiple phrases and hence prone to misinterpretation. 
(iii) Word level LDP frameworks. The word-level LDP frameworks \cite{lyu2020towards, plant2021cape, chen2023customized, tong2023privinfer, feyisetan2020privacy, xu2020differentially, carvalho2023tem, yue2021, zhou2023textobfuscator} perturb words or tokens independently. This can lead to a lack of semantic coherence in the generated text. Consequently, these approaches may fail to provide effective context for guiding LLMs, especially for text-generation tasks. Note that these methods were originally developed for privacy preserving classification and their effectiveness on privacy preserving LLM prompting needs more research. 
(iv) DP synthetic data generation frameworks. For DP synthetic data generation frameworks \cite{yue2022, flemings2024, kurakin2023, carranza2023}, the synthetic data generated by the DP fine-tuned models captures the general statistics of the private data, but it does not replicate all the details. This implies that while DP safeguards the privacy of individual samples in the original text, it also inhibits the model from learning the tails of the training distribution, thereby hindering the generation of rare patterns in the synthetic dataset, thus affecting semantic features. 
(v) PATE-based frameworks. The PATE-based frameworks \cite{duan2023flocks, tian2022seqpate} rely on the teacher-student framework. Knowledge distillation from teacher models to a student model can work well for classification tasks. However, for text generation tasks, the distilled knowledge may fail to fit in the context of the public dataset the student model accesses, leading to semantically inaccurate generated texts.  
 
\subsubsection{Privacy Challenges}
While significant efforts have been dedicated to addressing the privacy challenge, imperfections persist in achieving perfect privacy. We highlight some weaknesses categorized by privacy techniques and present them as follows: (i) Sanitization-based frameworks. Sanitization-based frameworks \cite{kan2023, chen2023, zhang2024cogenesis} necessitate the identification of sensitive attributes for anonymization. However, the sensitivity of certain attributes is domain and context dependent. For instance, terms related to sexual orientation such as transgender or bisexual may be deemed sensitive when discussing an individual's sexual orientation but non-sensitive in the context of general LGBTQ+ discussions. Consequently, this presents a gap in achieving comprehensive privacy. (ii) word-level perturbation frameworks. Although the effectiveness of the word-level perturbation framework, TextObfuscator \cite{zhou2023textobfuscator} has been examined experimentally, it lacks a rigorous mathematical proof and does not provide privacy guarantee. (iii) Document-level perturbation frameworks. Similarly, much as the document-level framework, DP-Prompt \cite{utpala2023locally} can conceal the authors' writing style, there remains a potential risk of inadvertently revealing personal information such as zip codes, bank details, gender, etc., when an LLM is prompted without due caution. 
(iv) Demonstration examples at the LLM. The DP-ICL \cite{wu2023privacy}  assumes the existence of demonstration examples at the LLM and suffers from a limited number of queries. With no privacy accounting technique, the privacy budget can potentially be exhausted if the number of queries exceeds a certain limit, thus a privacy risk. 

\subsubsection{Trustworthiness Challenges}
Apart from the aforementioned privacy challenges, issues associated with other trustworthiness aspects also exist. We present them as follows: (i) Server security. The lattice-based framework, LatticeGen \cite{zhang2023latticegen} and the cryptographic-based frameworks \cite{hou2023ciphergpt, hao2022iron, chen2022thex} that collaboratively generate tokens, share the generation control between the LLM server and the user. Granting users control over token selection and generation can compromise server security. For instance, in the LatticeGen framework, users have the authority to choose tokens during the generation process. This user privilege could potentially lead to jailbreaking attacks on the server if maliciously exploited.
(ii) Fairness in generated data. The conditional generation of DP synthetic data using DP synthetic data generation frameworks \cite{yue2022, flemings2024, kurakin2023, carranza2023} can disproportionately impact classes of varying sizes. Specifically, tight DP guarantees adversely affect learning the distribution of small-sized classes, leading to the models consistently generating large-sized classes. Thus resulting in unfairness in the generated DP synthetic data.

\subsection{Future Prospects}
\subsubsection{Computationally Efficient Private Prompting}
Although current private ICL methods have demonstrated promising results, many of them suffer from computational inefficiency. Further research is needed to address this challenge. We outline the key prospects as follows: (i) Perturbation and sanitization without a local LLM. Several works \cite{kan2023, chen2023, zhang2024cogenesis, tang2023, hong2023, duan2023flocks} require the assistance of a local LLM for DP noise addition and sanitization of sensitive attributes in users' texts. In reality, everyday users are likely to use computational devices with limited capabilities such as mobile phones, office laptops, etc., in executing their tasks. Hence, there is a need to devise new methods for efficiently perturbing and sanitizing private data during prompting, eliminating the necessity for a local LLM. 
(ii) Ensembling efficiency. Protecting privacy by ensembling during prompting requires sending multiple subsets of demonstration examples and/or a query to the LLM. Determining an appropriate number of example subsets is an interesting future direction for minimizing computational demands.
(iii) Universal fine-tuning for lattice configurations. Exploring the development of a unified format for linearizing lattices, which a single LLM can process across various lattice configurations, is worth investigating to scale down the fine-tuning requirements for each lattice configuration in the LatticeGen \cite{zhang2023latticegen} framework.
(iv) Cryptographic efficiency. The current implementation of CipherGPT \cite{hou2023ciphergpt} uses a single thread. Leveraging parallel computing technologies such as GPU and FPGA can speed up their execution.  Furthermore, computing architectures such as in-memory and in-storage can be explored to boost the speed. An alternative direction to explore for cryptographic frameworks \cite{hou2023ciphergpt, hao2022iron} is to modify the model structure to be more crypto-friendly. Reducing the number of activations as used in DeepSecure \cite{rouhani2018deepsecure} can be explored to create crypto-friendly frameworks.

\subsubsection{Mitigating Semantic Inadequacy}
Several private prompting techniques suffer from semantic inaccuracies. Further investigations are required to address this challenge. We present the main concepts that require further investigations to tackle this issue as follows: (i) Obfuscation/lattice-based frameworks. The poor text quality issue in the lattice-based framework LatticeGen \cite{zhang2023latticegen} can be mitigated by employing larger m-gram units. However, this strategy leads to an exponential increase in inference computation as inference is run on an exponential number of options. In the future, exploring an approach to strategically select small portions can be pursued. Furthermore, adapting the obfuscation-based framework IOI \cite{yao2024} for text generation tasks can be further investigated. Crucially, resolving issues such as mix-up tokens and variable lengths of generated texts is necessary for text generation tasks. 
(ii) Encryption-based frameworks. Due to emoji size vs text size mismatch, the emoji-based framework EmojiCrypt \cite{lin2024} can lead to emoji misinterpretation problems. Further work to address these problems can be conducted. A solution can evolve around expanding the emoji space to match the text space. An alternative approach could involve developing models capable of mapping the limited emoji space to text space and vice versa, depending on context.
(iii) Further explorations are equally required to improve the generated text quality in PATE-based, word-level perturbation, and DP synthetic data generation frameworks.

\subsubsection{Improving Privacy Protection}
The privacy preservation capabilities of a number of the frameworks can be enhanced. We organize the possible future directions towards enhancing privacy as follows: (i) Sanitization-based frameworks. The sanitization-based frameworks \cite{kan2023, chen2023, zhang2024cogenesis} suffer from domain and context sensitivity as stated in section \ref{sec:lim}. Thus, investigating the development of robust, lightweight models capable of identifying sensitive attributes based on their domain context can be explored as a solution to this challenge in the future.
(ii) Sentence/document-level perturbation frameworks. From Table \ref{tab:prompt_eng_dp}, we can see that privacy protection in the sentence /document levels is still lacking in studies. Word-level perturbation approaches do not consider sentence or document level privacy which is important and practical in the NLP scenario. We should not only consider the privacy protection of each word but also consider how to hide sentence level secret information. Thus designing sentence and document level private prompting is an important but challenging problem. We should also consider privacy information at multiple levels by integrating different techniques. For example, to safeguard PII in document-level perturbation framework DP-Prompt \cite{utpala2023locally}, one can define a set of sensitive attributes and prompt the LLM to replace these attributes with a dummy identifier while paraphrasing can be explored. Further investigation is also warranted to explore the impact of different prompt templates and hallucinations on the paraphrases within the context of the privacy-utility tradeoff. 
(iii) Soft prompts. Another line of work can focus on privacy risk analysis for soft prompts. While significant research has been devoted to mitigating privacy risks through ensembling with discrete prompts \cite{duan2023}, little attention has been directed towards soft prompts. There remains an opportunity to further explore the privacy implications of soft prompts and potential mitigation strategies.  
(iv) External datastore. Improving a user's prompt through incorporating external information has been mooted. It is imperative to develop methods for protecting the privacy of retrieved information from external datastores. Adopting mechanisms with proven privacy guarantees such as DP, homomorphic encryption, etc., to privately retrieve information from external datastores appears to be an interesting direction.
(v) Privacy-utility trade-off comparison with fine-tuning. Several studies \cite{pecher2024fine, sun2024fine, mosbach2023few} have endeavored to compare prompting-based methods with fine-tuning across factors such as required training data volume, comprehension of human values, and computational requirements. However, to date, there has been no research that compares the trade-off between privacy and utility specifically for model fine-tuning and prompting methods. Subsequent studies could delve into this direction for further investigation.

\subsubsection{Trustworthy Prompt Engineering} As shown in  \cite{sun2024trustllm}, there are eight facets of LLM trustworthiness, truthfulness, safety, fairness, robustness, privacy, machine ethics, transparency, and accountability. Our survey focuses on privacy protection in prompting. There has been very few studies on how the developed privacy preserving prompting frameworks would impact other facets of trustworthiness.  The collaborative LatticeGen \cite{zhang2023latticegen} and cryptographic-based \cite{hou2023ciphergpt, hao2022iron, chen2022thex} frameworks can compromise the security of the server. To address this challenge, it is essential to further conduct an analysis regarding the security risks posed to the server by sharing generation control with the user. Furthermore, DP has a detrimental effect on small-sized classes, resulting in unfairness in synthetic data generation frameworks \cite{yue2022, flemings2024, kurakin2023, carranza2023}. 
Future studies are greatly needed to investigate the impact of privacy preserving prompting on LLM's performance from other trustworthiness facets including truthfulness, safety, fairness, and robustness, and develop trustworthy prompt engineering. 

\subsubsection{Extension to Other Modalities}
Vision-Language Models (VLMs) have been intensively investigated recently\cite{zhang2024vision}. VLMs learn rich vision-language correlation from web-scale image-text pairs and enable zero-shot and few-shot predictions on various visual recognition tasks. However, the majority of existing research on privacy preserving prompt engineering focuses on text and tabular data analysis tasks. It is interesting and imperative to study privacy protection prompting with VLMs. Ideas from some approaches covered in this survey could be adapted to these new scenarios. For example, DP-forward \cite{du2023} has showcased its capability to uphold privacy during inference by introducing DP noise to the text embedding space in the forward pass. In this case, users only need to download specific pipeline components to generate the noisy embeddings. Investigating the extension of such methodologies to transformer-based foundational models for vision, audio, and video could be a promising avenue for future research. 

\subsubsection{Benchmark Datasets and Open Source Software} 
Resources are still limited to spur further research and development in this area. For example, most research in privacy-preserving prompting, with focuses on NLP tasks like creative writing, information extraction, and question-answering, rely on synthetic data for evaluation. However, the availability of standardized benchmark real-world datasets for studying these privacy-preserving models are limited.  Establishing evaluation benchmarks for privacy-preserving prompting in these NLP tasks would facilitate consistent and measurable progress in the field. Additionally, to promote the integration of privacy-preserving prompting frameworks into real-world systems and to provide frameworks for the research community, it is essential to develop open-source libraries encompassing proven privacy mechanisms for dedicated privacy-preserving prompting and ICL. The aforementioned software tools primarily rely on anonymization mechanisms, rendering them susceptible to the weaknesses inherent in anonymization mechanisms \cite{rubinstein2016anonymization}. Integrating mechanisms with established privacy assurances into these software tools can empower domain users and developers to seamlessly integrate the libraries into their systems or devise customized privacy-preserving frameworks using the provided APIs. 

\section{Conclusion}\label{sec:conclusion}
LLMs have garnered substantial attention from both industry and academia recently.
Their standout feature lies in their capability to make predictions when provided with instruction and/or demonstration examples.
However, numerous studies have illustrated how malicious entities can exploit this ability to breach privacy, encouraging the development of various frameworks aimed at mitigating this challenge. In this survey, we comprehensively examined the frameworks designed to safeguard privacy during ICL specifically, as well as prompting in general. We have systematically structured these frameworks according to the privacy mechanisms they employ. Additionally, we have established connections between the different frameworks based on their respective privacy objectives and methodologies. Furthermore, we provided an overview of the common resources utilized for developing and evaluating privacy-preserving prompting systems.
 We extensively discussed the limitations inherent in existing works and identified promising areas necessitating further investigation. We aspire for this to stimulate increased interest and advancement in the field of privacy-preserving prompting.

\section*{Acknowledgements}
This work was supported in part by the National Science Foundation under awards 1920920 and 1946391.

\ifCLASSOPTIONcaptionsoff
  \newpage
\fi

\bibliographystyle{IEEEtran}
\bibliography{ref}

\end{document}